\documentclass[10pt,twocolumn,letterpaper]{article}

\usepackage{cvpr}
\usepackage{times}
\usepackage{epsfig}
\usepackage{graphicx}
\usepackage{amsmath}
\usepackage{amssymb}
\usepackage{bm}
\usepackage{array}
\usepackage{booktabs}
\usepackage{authblk}

\usepackage{color}

\newcommand{\Tref}[1]{Table~\ref{#1}}

\newcommand{\fref}[1]{Fig.~\ref{#1}}
\newcommand{\Fref}[1]{Figure~\ref{#1}}
\newcommand{\sref}[1]{Sec.~\ref{#1}}

\def\rnum#1{\expandafter{\romannumeral #1}}

\usepackage[breaklinks=true,bookmarks=false]{hyperref}

\cvprfinalcopy 

\begin{document}

\makeatletter
\newcommand{\argmax}{\mathop{\rm arg~max}\limits}
\newcommand{\tblcaption}[1]{\def\@captype{table}\caption{#1}}
\makeatother

\title{RotationNet: Joint Object Categorization and Pose Estimation \\ Using Multiviews from Unsupervised Viewpoints}

\author[1]{Asako Kanezaki\thanks{kanezaki.asako@aist.go.jp}}
\author[2]{Yasuyuki Matsushita\thanks{yasumat@ist.osaka-u.ac.jp}}
\author[1]{Yoshifumi Nishida\thanks{y.nishida@aist.go.jp}}
\affil[1]{National Institute of Advanced Industrial Science and Technology (AIST)}
\affil[2]{Graduate School of Information Science and Technology, Osaka University}

\renewcommand\Authands{ and }

\maketitle

\begin{abstract}
  We propose a Convolutional Neural Network (CNN)-based model ``RotationNet,'' which takes multi-view images of an object as input and jointly estimates its pose and object category. Unlike previous approaches that use known viewpoint labels for training, our method treats the viewpoint labels as latent variables, which are learned in an unsupervised manner during the training using an unaligned object dataset. RotationNet is designed to use only a partial set of multi-view images for inference, and this property makes it useful in practical scenarios where only partial views are available. 
  Moreover, our pose alignment strategy enables one to obtain view-specific feature representations shared across classes, which is important to maintain high accuracy in both object categorization and pose estimation.
  Effectiveness of RotationNet is demonstrated by its superior performance to the state-of-the-art methods of 3D object classification on $10$- and $40$-class ModelNet datasets. 
  We also show that RotationNet, even trained without known poses, achieves the state-of-the-art performance on an object pose estimation dataset.
  The code is available on \url{https://github.com/kanezaki/rotationnet}

\end{abstract}

\section{Introduction}

Object classification accuracy can be enhanced by the use of multiple different views of a target object~\cite{borotschnig2000appearance,paletta2000active}. Recent remarkable advances in image recognition and collection of 3D object models enabled the learning of multi-view representations of objects in various categories. However, in real-world scenarios, objects can often only be observed from limited viewpoints due to occlusions, which makes it difficult to rely on multi-view representations that are learned with the whole circumference. The desired property for the real-world object classification is that, when a viewer observes a partial set ($ \geq 1$ images) of the full multi-view images of an object, it should recognize from which directions it observed the target object to correctly infer the category of the object.
It has been understood that if the viewpoint is known the object classification accuracy can be improved. Likewise, if the object category is known, that helps infer the viewpoint. As such, object classification and viewpoint estimation is a tightly coupled problem, which can best benefit from their joint estimation.

\begin{figure}[t]
  \begin{center}
    \includegraphics[width=.9\linewidth]{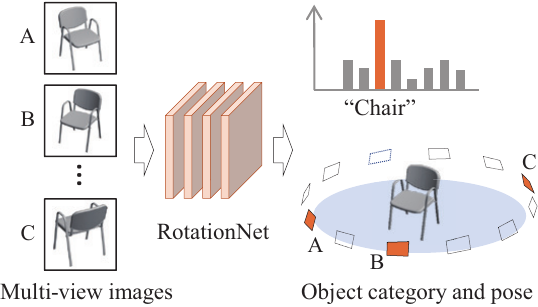}
  \end{center}
  \vspace{-3mm}
  \caption{Illustration of the proposed method \textit{RotationNet}.
    RotationNet takes a partial set ($ \geq 1$ images) of the full multi-view images of an object as input and predicts its object category by rotation, where the best pose is selected to maximize the object category likelihood. Here, viewpoints from which the images are observed are jointly estimated to predict the pose of the object.
  }
  \label{fig:overview}
\end{figure}

We propose a new Convolutional Neural Network (CNN) model that we call \textit{RotationNet}, which takes multi-view images of an object as input and predicts its pose and object category (\fref{fig:overview}). RotationNet outputs viewpoint-specific category likelihoods corresponding to all pre-defined discrete viewpoints for each image input, and then selects the object pose that maximizes the integrated object category likelihood. Whereas, at the training phase, RotationNet uses a complete set of multi-view images of an object captured from all the pre-defined viewpoints, for inference it is able to work with only a partial set of all the multi-view images -- a single image at minimum -- as input. 
In addition, RotationNet does not require the multi-view images to be provided at once but allows their sequential input and updates of the target object's category likelihood. This property is suitable for applications that require on-the-fly classification with a moving camera. 

\begin{figure*}[t]
  \begin{center}
    \includegraphics[width=\textwidth]{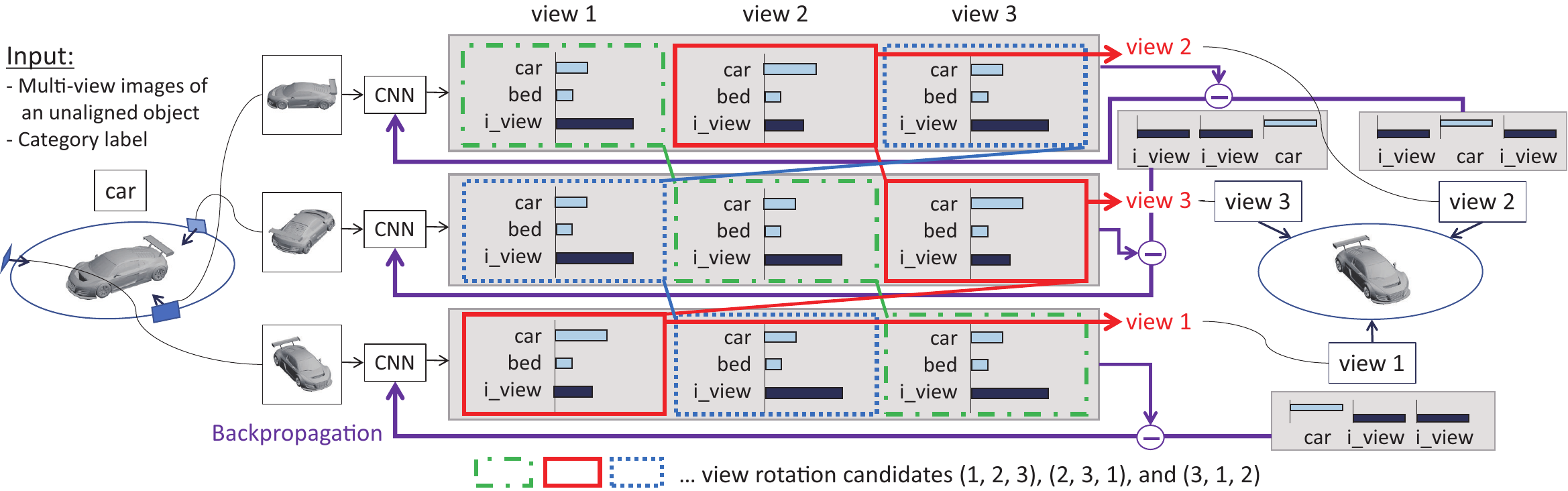}
  \end{center}
  \vspace{-3mm}
  \caption{Illustration of the training process of RotationNet, where the number of views $M$ is 3 and the number of categories $N$ is 2. A training sample consists of $M$ images of an unaligned object and its category label $y$. For each input image, our CNN (RotationNet) outputs $M$ histograms with $N+1$ bins whose norm is 1. The last bin of each histogram represents the ``incorrect view'' class, which serves as a weight of how likely the histogram does not correspond to each viewpoint variable. According to the histogram values, we decide which image corresponds to views 1, 2, and 3. There are three candidates for view rotation: (1, 2, 3), (2, 3, 1), and (3, 1, 2). For each candidate, we calculate the score for the ground-truth category (``car'' in this case) by multiplying the histograms and selecting the best choice: (2, 3, 1) in this case. Finally, we update the CNN parameters in a standard back-propagation manner with the estimated viewpoint variables. 
  Note that it is the same CNN that is being used.}
  \label{fig:rotationnet}
\end{figure*}

The most representative feature of RotationNet is that it treats viewpoints where training images are observed as \emph{latent} variables during the training (\fref{fig:rotationnet}). This enables unsupervised learning of object poses using an unaligned object dataset; thus, it eliminates the need of preprocessing for pose normalization that is often sensitive to noise and individual differences in shape. Our method automatically determines the basis axes of objects based on their appearance during the training and achieves not only intra-class but also inter-class object pose alignment. Inter-class pose alignment is important to deal with joint learning of object pose and category, because the importance of object classification lies in emphasizing differences in different categories when their appearances are similar. Without inter-class pose alignment, it may become an ill-posed problem to obtain a model to distinguish, \eg, a car and a bus if the side view of a car is compared with the frontal view of a bus.

Our main contributions are described as follows. We first show that RotationNet outperforms the current state-of-the-art classification performance on 3D object benchmark datasets consisting of $10$- and $40$-categories by a large margin (\Tref{table:classification}). Next, even though it is trained without the ground-truth poses, RotationNet achieves superior performance to previous works on an object pose estimation dataset. We also show that our model generalizes well to a real-world image dataset that was newly created for the general task of multi-view object classification.
Finally, we train RotationNet with the new dataset named MIRO and demonstrate the performance of real-world applications using a moving USB camera or a head-mounted camera (Microsoft HoloLens).

\section{Related work}
\vspace{-1.3mm}
There are two main approaches for the CNN-based 3D object classification: voxel-based and 2D image-based approaches. The earliest work on the former approach is 3D ShapeNets~\cite{wu20153d}, which learns a Convolutional Deep Belief Network that outputs probability distributions of binary occupancy voxel values.
Latest works on similar approaches showcased improved performance~\cite{maturana2015voxnet,li_fpnn2016,Jiajun2016}.
Even when working with 3D objects, 2D image-based approaches are shown effective for general object recognition tasks.
Su~\etal~\cite{su15mvcnn} proposed multi-view CNN (MVCNN), which takes multi-view images of an object captured from surrounding virtual cameras as input and outputs the object's category label.
Multi-view representations are also used for 3D shape retrieval~\cite{bai2016gift}. 
Qi~\etal~\cite{Qi_2016_CVPR} gives a comprehensive study on the voxel-based CNNs and multi-view CNNs for 3D object classification.
Other than those above, point-based approach~\cite{garcia2016pointnet,qi2016pointnet,klokov2017} is recently drawing much attention; however, the performance on 3D object classification is yet inferior to those of multi-view approaches.
The current state-of-the-art result on the ModelNet40 benchmark dataset is reported by Wang~\etal~\cite{wang2017bmvc}, which is also based on the multi-view approach.

Because MVCNN integrates multi-views in a view-pooling layer which lies in the middle of the CNN, it requires a complete set of multi-view images recorded from all the pre-defined viewpoints for object inference.
Unlike MVCNN, our method is able to classify an object using a \textit{partial} set of multi-view images that may be sequentially observed by a moving camera. 
Elhoseiny~\etal~\cite{elhoseiny2016comparative} explored CNN architectures for joint object classification and pose estimation learned with multi-view images.
Whereas their method takes a single image as input for its prediction, we mainly focus on how to aggregate predictions from multiple images captured from different viewpoints.


Viewpoint estimation is significant in its role in improving object classification.
Better performance was achieved on face identification~\cite{NIPS2014_5546}, human action classification~\cite{chen2014inferring}, and image retrieval~\cite{su20153d} by generating unseen views after observing a single view.
These methods ``imagine'' the appearance of objects' unobserved profiles, which is innately more uncertain than using real observations. 
Sedaghat~\etal~\cite{sedaghat2016orientation} proposed a voxel-based CNN that outputs orientation labels as well as classification labels and demonstrated that it improved 3D object classification performance.

All the methods mentioned above assume known poses in training samples; however, object poses are not always aligned in existing object databases.
Novotny~\etal~\cite{Novotny_2017_ICCV} proposed a viewpoint factorization network that utilizes relative pose changes within each sequence to align objects in videos in an unsupervised manner.
Our method also aligns object poses via unsupervised viewpoint estimation, where viewpoints of images are treated as \emph{latent} variables during the training.
Here, viewpoint estimation is learned in an unsupervised manner to best promote the object categorization task.
In such a perspective, our method is related to
Zhou~\etal~\cite{zhou2017unsupervised}, where view synthesis is trained as the ``meta''-task to train multi-view pose networks by utilizing the synthesized views as the supervisory signal.

Although joint learning of object classification and pose estimation has been widely studied~\cite{savarese20073d,lai_aaai11,zhang2013joint,bakry2014untangling,Su_2015_ICCV}, inter-class pose alignment has drawn little attention.
However, it is beneficial to share view-specific appearance information across classes to simultaneously solve for object classification and pose estimation.
Kuznetsova~\etal~\cite{Kuznetsova2016} pointed out this issue and presented a metric learning approach that shares visual components across categories for simultaneous pose estimation and class prediction.
Our method also uses a model with view-specific appearances that are shared across classes; thus, it is able to maintain high accuracy for both object classification and pose estimation.

\section{Proposed method}

The training process of RotationNet is illustrated in \fref{fig:rotationnet}.
We assume that multi-view images of each training object instance are observed from all the pre-defined viewpoints.
Let $M$ be the number of the pre-defined viewpoints and $N$ denote the number of target object categories.
A training sample consists of $M$ images of an object $\{{\bm x}_i\}_{i=1}^M$ and its category label $y \in \{1,\dots,N\}$.
We attach a viewpoint variable $v_i \in \{1,\dots,M\}$ to each image ${\bm x}_i$ and set it to $j$ when the image is observed from the $j$-th viewpoint, \ie, $v_i \leftarrow j$. 
In our method, only the category label $y$ is given during the training whereas the viewpoint variables $\{v_i\}$ are \textit{unknown}, namely, $\{v_i\}$ are \textbf{treated as latent variables that are optimized in the training process}.

RotationNet is defined as a differentiable multi-layer neural network $R( \cdot )$.
The final layer of RotationNet is the concatenation of $M$ softmax layers, each of which outputs the category likelihood \hbox{$P( \hat{y}_i \mid {\bm x}_i, v_i=j )$} where $j \in \{1,\ldots,M\}$ for each image ${\bm x}_i$.
Here, $\hat{y}_i$ denotes an estimate of the object category label for ${\bm x}_i$.
For the training of RotationNet, we input the set of images $\{{\bm x}_i\}_{i=1}^M$ simultaneously and solve the following optimization problem:
\begin{equation}
  \max_{R,\{v_i\}_{i=1}^M} \prod_{i=1}^M P( \hat{y}_i = y \mid {\bm x}_i, v_i ). \label{eq:rotationnet}
\end{equation}
The parameters of $R$ and latent variables $\{v_i\}_{i=1}^M$ are optimized to output the highest probability of $y$ for the input of multi-view images $\{{\bm x}_i\}_{i=1}^M$.

Now, we describe how we design $P( \hat{y}_i \mid {\bm x}_i, v_i )$ outputs.
First of all, the category likelihood $P( \hat{y}_i = y \mid {\bm x}_i, v_i )$ should become close to one when the estimated $v_i$ is correct; in other words, the image ${\bm x}_i$ is truly captured from the $v_i$-th viewpoint.
Otherwise, in the case that the estimated $v_i$ is incorrect, $P( \hat{y}_i = y \mid {\bm x}_i, v_i )$ may not necessarily be high because the image ${\bm x}_i$ is captured from a different viewpoint.
As described above, we decide the viewpoint variables $\{v_i\}_{i=1}^M$ according to the $P( \hat{y}_i = y \mid {\bm x}_i, v_i )$ outputs as in (\ref{eq:rotationnet}).
In order to obtain a stable solution of $\{v_i\}_{i=1}^M$ in (\ref{eq:rotationnet}), we introduce an ``incorrect view'' class and append it to the target category classes.
Here, the ``incorrect view'' class plays a similar role to the ``background'' class for object detection tasks, which represents negative samples that belong to a ``non-target'' class.
Then, RotationNet calculates $P( \hat{y}_i \mid {\bm x}_i, v_i )$ by applying softmax functions to the $(N+1)$-dimensional outputs, where $\sum_{\hat{y}_i = 1}^{N+1} P( \hat{y}_i \mid {\bm x}_i, v_i ) = 1$.
Note that $P\left( \hat{y}_i=N+1 \mid {\bm x}_i, v_i \right)$, which corresponds to the probability that the image ${\bm x}_i$ belongs to the ``incorrect view'' class for the $v_i$-th viewpoint, indicates how likely it is that the estimated viewpoint variable $v_i$ is incorrect.
   
Based on the above discussion, we substantiate (\ref{eq:rotationnet}) as follows.
Letting $P_i = \left[ p^{(i)}_{j,k} \right] \in \mathbb{R}_+^{M \times (N+1)} $ denote a matrix composed of $P( \hat{y}_i \mid {\bm x}_i, v_i )$ for all the $M$ viewpoints and $N+1$ classes, the target value of $P_i$ in the case that $v_i$ is correctly estimated is defined as follows:
\begin{equation}
  p^{(i)}_{j,k} = 
  \begin{cases}
    1 & \!\!\! (j=v_i \hspace{1mm}{\rm and}\hspace{1mm} k = y)~\mathrm{or}~
     (j \neq v_i \hspace{1mm}{\rm and}\hspace{1mm} k = N+1) \\
    0 & \!\!\! (\mathrm{otherwise}).
  \end{cases}
\end{equation}
In this way, (\ref{eq:rotationnet}) can be rewritten as the following cross-entropy optimization problem: 
\begin{equation}
  \max_{R,\{v_i\}_{i=1}^M} \sum_{i=1}^M \left( \log p^{(i)}_{v_i,y} + \sum_{j \neq v_i}  \log p^{(i)}_{j,N+1}  \right).
\end{equation}
If we fix $\{v_i\}_{i=1}^M$ here, the above can be written as a subproblem of optimizing $R$ as follows:
\begin{equation}
  \max_{R} \sum_{i=1}^M \left( \log p^{(i)}_{v_i,y} + \sum_{j \neq v_i}  \log p^{(i)}_{j,N+1}  \right), \label{eq:fixedvi}
\end{equation}
where the parameters of $R$ can be iteratively updated via standard back-propagation of $M$ softmax losses. 
Since $\{v_i\}_{i=1}^M$ are not constant but latent variables that need to be optimized during the training of $R$,
we employ alternating optimization of $R$ and $\{v_i\}_{i=1}^M$.
More specifically, in every iteration, our method determines $\{v_i\}_{i=1}^M$ according to $P_i$ obtained via forwarding of (fixed) $R$, and then update $R$ according to the estimated $\{v_i\}_{i=1}^M$ by fixing them.

The latent viewpoint variables $\{v_i\}_{i=1}^M$ are determined by solving the following problem:
\begin{eqnarray}
\hspace{-3mm} &&  \hspace{-3mm} \max_{\{v_i\}_{i=1}^M} \sum_{i=1}^M \left( \log p^{(i)}_{v_i,y} + \sum_{j \neq v_i}  \log p^{(i)}_{j,N+1}  \right) \nonumber \\
\hspace{-3mm} &=& \hspace{-3mm} \max_{\{v_i\}_{i=1}^M} \sum_{i=1}^M \left( \log p^{(i)}_{v_i,y} + \sum_{j=1}^M \log p^{(i)}_{j,N+1} - \log p^{(i)}_{v_i,N+1}  \right) \nonumber \\
\hspace{-3mm} &=& \hspace{-3mm} \max_{\{v_i\}_{i=1}^M} \prod_{i=1}^M \frac{ p^{(i)}_{v_i,y} }{ p^{(i)}_{v_i,N+1}  }, \label{eq:v_i}
\end{eqnarray}
%
in which the conversion used the fact that $\sum_{j=1}^M \log p^{(i)}_{j,N+1}$ is constant w.r.t. $\{v_i\}_{i=1}^M$.
Because the number of candidates for $\{v_i\}_{i=1}^M$ is limited, we calculate the evaluation value of (\ref{eq:v_i}) for all the candidates and take the best choice.
The decision of $\{v_i\}_{i=1}^M$ in this way emphasizes view-specific features for object categorization, which contributes to the self-alignment of objects in the dataset.

In the inference phase, RotationNet takes as input $M'$ $(1 \leq M' \leq M)$ images of a test object instance, either simultaneously or sequentially, and outputs $M'$ probabilities. Finally, it integrates the $M'$ outputs to estimate the category of the object and the viewpoint variables as follows:
\begin{equation}
  \left\{\hat{y}, \{\hat{v}_i \}_{i=1}^{M'}\right\} = \argmax_{y,\{v_i\}_{i=1}^{M'}} \prod_{i=1}^{M'} \frac{ p^{(i)}_{v_i,y} }{ p^{(i)}_{v_i,N+1}  }.
\end{equation}
Similarly to the training phase, we decide $\{\hat{v}_i\}_{i=1}^{M'}$ according to the outputs $\{P_i\}_{i=1}^{M'}$.
Thus RotationNet is able to estimate the pose of the object as well as its category label.

\begin{figure}[t]
  \begin{center}
    \begin{minipage}{.26\hsize}
      \includegraphics[width=\linewidth]{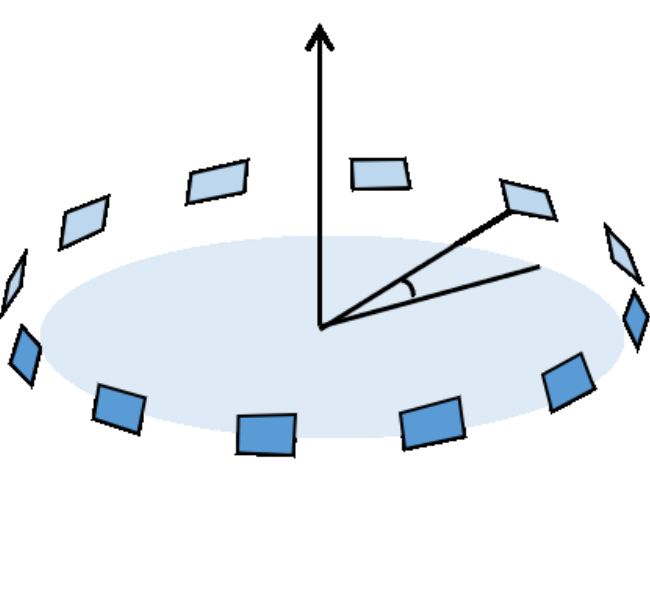}
      \centering
      case (\rnum{1})
    \end{minipage}
    \hspace{1mm}
    \begin{minipage}{.25\hsize}
      \includegraphics[width=\linewidth]{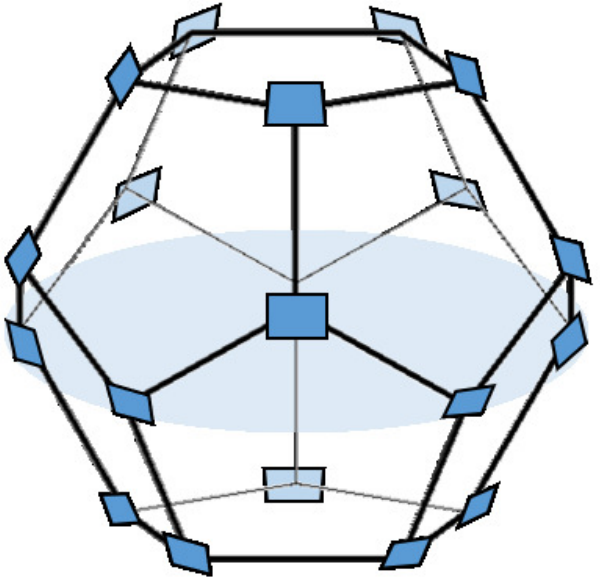}
      \centering
      case (\rnum{2})
    \end{minipage}      
    \hspace{1mm}
    \begin{minipage}{.27\hsize}
      \includegraphics[width=\linewidth]{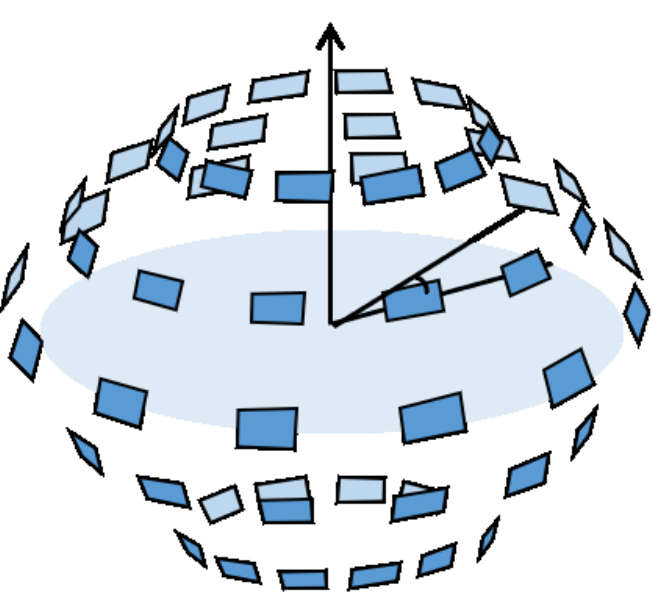}
      \centering
      case (\rnum{3})
    \end{minipage}      
  \end{center}
  \vspace{-2mm}
  \caption{Illustration of three viewpoint setups considered in this work. A target object is placed on the center of each circle.}
  \label{fig:viewpoints_settings}
\end{figure}

\paragraph{Viewpoint setups for training}
While choices of the viewpoint variables $\{v_i\}_{i=1}^{M'}$ can be arbitrary, we consider two setups in this paper, with and without an upright orientation assumption, similarly to MVCNN~\cite{su15mvcnn}. The former case is often useful with images of real objects captured with one-dimensional turning tables, whereas the latter case is rather suitable for unaligned 3D models. 
We also consider the third case that is also based on the upright orientation assumption (as the first case) but with multiple elevation levels.
We illustrate the three viewpoint setups in \fref{fig:viewpoints_settings}.

\noindent{\bf Case (i): with upright orientation}~~
In the case where we assume upright orientation, we fix a specific axis as the rotation axis (\eg, the $z$-axis), which defines the upright orientation, and then place viewpoints at intervals of the angle $\theta$ around the axis, elevated by $\phi$ (set to $30^\circ$ in this paper) from the ground plane.
We set $\theta=30^\circ$ in default, which yields $12$ views for an object ($M=12$).
We define that ``view $m+1$'' is obtained by rotating the view position ``view $m$'' by the angle $\theta$ about the $z$-axis.
Note that the view obtained by rotating ``view $M$'' by the angle $\theta$ about the $z$-axis corresponds to ``view 1.''
We assume the sequence of input images is consistent with respect to a certain direction of rotation in the training phase.
For instance, if $v_i$ is $m$ $(m<M)$, then $v_{i+1}$ is $m+1$.
Thus the number of candidates for all the viewpoint variables $\{v_i\}_{i=1}^M$ is $M$.

\noindent{\bf Case (ii): w/o upright orientation}~~
In the case where we do not assume upright orientation, we place virtual cameras on the $M=20$ vertices of a dodecahedron encompassing the object.
This is because a dodecahedron has the largest number of vertices among regular polyhedra, where viewpoints can be completely equally distributed in 3D space.
Unlike case (i), where there is a unique rotation direction, there are three different patterns of rotation from a certain view, because three edges are connected to each vertex of a dodecahedron.
Therefore, the number of candidates for all the viewpoint variables $\{v_i\}_{i=1}^M$ is $60$ ($= 3M$)\footnote{A dodecahedron has 60 orientation-preserving symmetries.}.

\noindent{\bf Case (\rnum{3}): with upright orientation and multiple elevation levels}~~
This case is an extension of case (i). Unlike case (i) where the elevation angle is fixed,
we place virtual cameras at intervals of $\phi$ in $[-90^\circ,90^\circ]$.
There are $M = M_a \times M_e$ viewpoints, where $M_a = \frac{360^\circ}{\theta}$ and $M_e = \frac{180^\circ}{\phi}+1$.
As with the case (i), the number of candidates for all the viewpoint variables $\{v_i\}_{i=1}^M$ is $M_a$ due to the upright orientation assumption.

\section{Experiments}
\label{sec:experiments}

In this section, we show the results of the experiments with 3D model benchmark datasets (\sref{subsec:3dmodel}), a real image benchmark dataset captured with a one-dimensional turning table  (\sref{subsec:1DOF}), and our new dataset consisting of multi-view real images of objects viewed with two rotational degrees of freedom (\sref{subsec:2DOF}).
%
The baseline architecture of our CNN is based on AlexNet~\cite{Krizhevsky_imagenetclassification}, which is smaller than the VGG-M network architecture that MVCNN~\cite{su15mvcnn} used.
To train RotationNet, we fine-tune the weights pre-trained using the ILSVRC 2012 dataset~\cite{ILSVRC15}.
We used classical momentum SGD with a learning rate of $0.0005$ and a momentum of $0.9$ for optimization.

As a baseline method, we also fine-tuned the pre-trained weights of a standard AlexNet CNN that only predicts object categories. 
To aggregate the predictions of multi-view images, we summed up all the scores obtained through the CNN. 
This method can be recognized as a modified version of MVCNN~\cite{su15mvcnn}, where the view-pooling layer is placed after the final softmax layer.
We chose average pooling for the view-pooling layer in this setting of the baseline, because we observed that the performance was better than that with max pooling.
We also implemented MVCNN~\cite{su15mvcnn} based on the AlexNet architecture with the original view-pooling layer for a fair comparison.

\subsection{Experiment on 3D model datasets}
\label{subsec:3dmodel}
\begin{figure*}[t]
  \begin{center}
    \begin{minipage}{.32\hsize}
      \includegraphics[width=\linewidth]{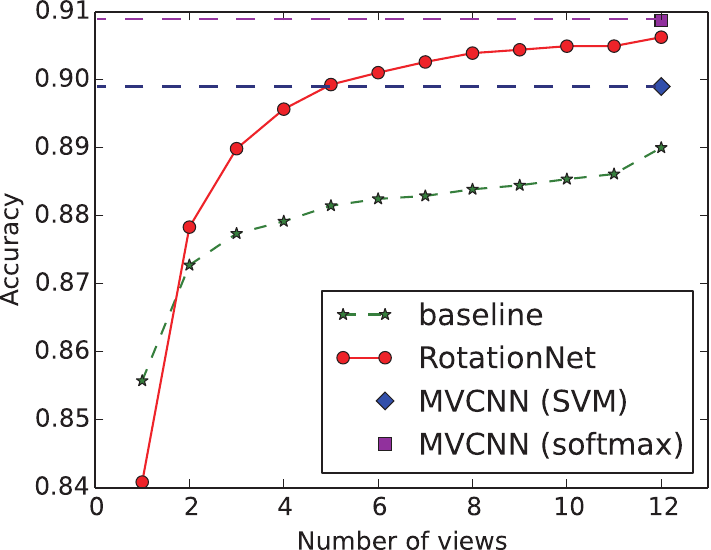}
      \centering
      (a) ModelNet40, case (i)
    \end{minipage}      
    \begin{minipage}{.32\hsize}
      \includegraphics[width=\linewidth]{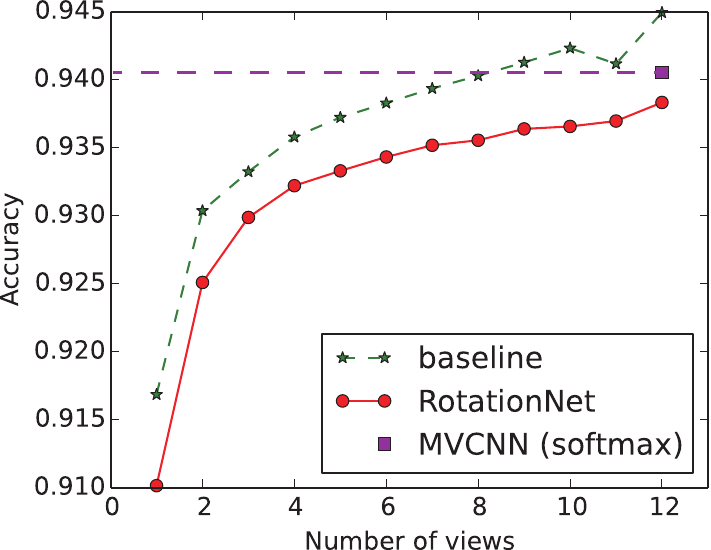}
      \centering
      (b) ModelNet10, case (i)
    \end{minipage}      
    \begin{minipage}{.32\hsize}
      \includegraphics[width=\linewidth]{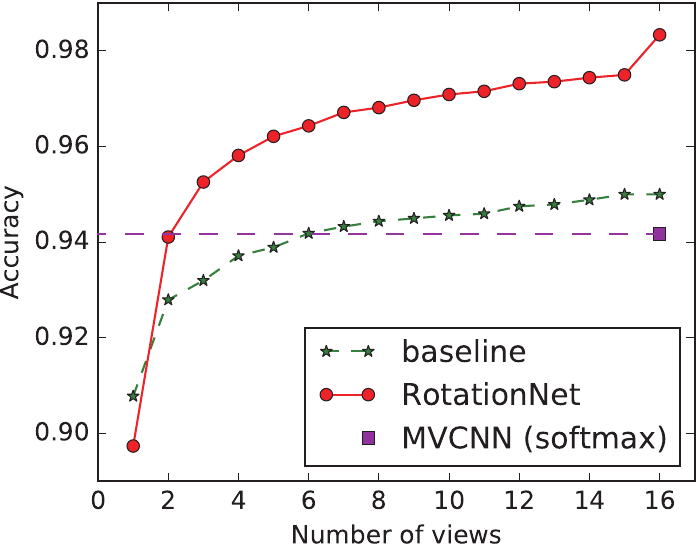}
      \centering
      (c) MIRO, case (i)
    \end{minipage}      

    \begin{minipage}{.32\hsize}
      \includegraphics[width=\linewidth]{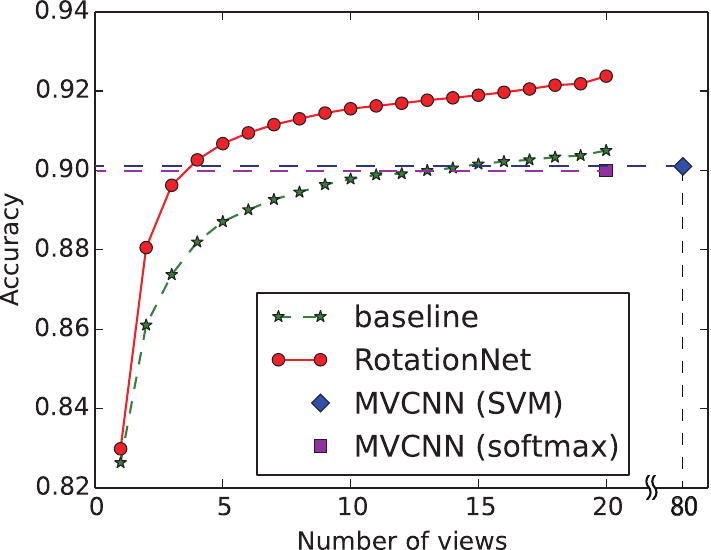}
      \centering
      (d) ModelNet40, case (ii)
    \end{minipage}      
    \begin{minipage}{.32\hsize}
      \includegraphics[width=\linewidth]{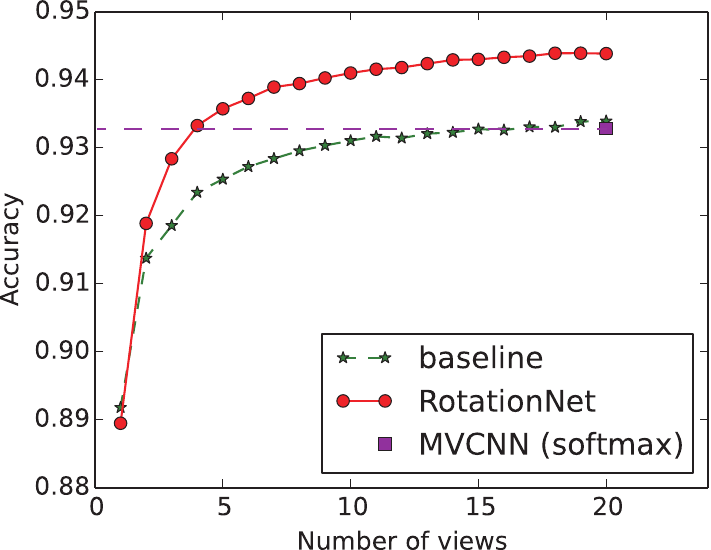}
      \centering
      (e) ModelNet10, case (ii)
    \end{minipage}      
    \begin{minipage}{.32\hsize}
      \includegraphics[width=\linewidth]{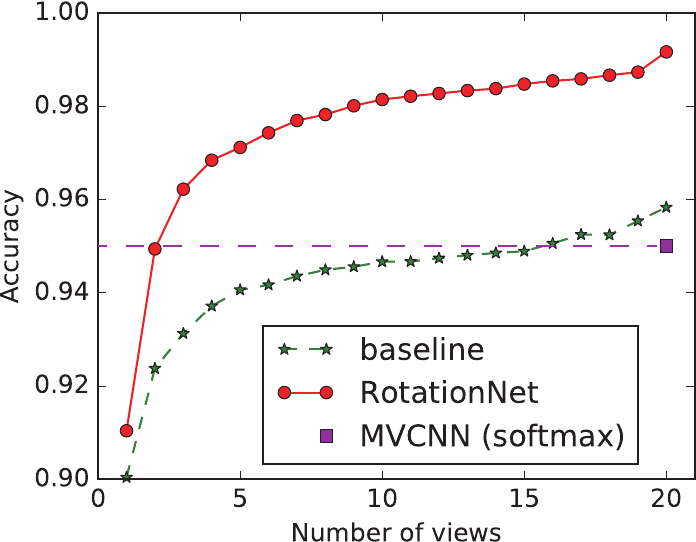}
      \centering
      (f) MIRO, case (ii)
    \end{minipage}      
  \end{center}
  \vspace{-1mm}
  \caption{Classification accuracy vs. number of views used for prediction. From left to right are shown the results on ModelNet40, ModelNet10, and our new dataset MIRO.
    The results in case (i) are shown in top and those in case (ii) are shown in bottom.
    See \Tref{table:classification} for an overall performance comparison to existing methods on ModelNet40 and ModelNet10.
  }
  \label{fig:results_case1_and_2}
\end{figure*}

We first describe the experimental results on two 3D model benchmark datasets, ModelNet10 and ModelNet40~\cite{wu20153d}.
ModelNet10 consists of 4,899 object instances in 10 categories, whereas ModelNet40 consists of 12,311 object instances in 40 categories.
First, we show the change of object classification accuracy versus the number of views used for prediction in cases (i) and (ii) with ModelNet40 and ModelNet10, respectively, in \fref{fig:results_case1_and_2} (a)-(b) and \fref{fig:results_case1_and_2} (d)-(e).
For fair comparison, we used the same training and test split of ModelNet40 as in~\cite{wu20153d} and~\cite{su15mvcnn}.
We prepared multi-view images (i) with the upright orientation assumption and (ii) without the upright orientation assumption using the rendering software published in~\cite{su15mvcnn}.
Here, we show the average scores of $120$ trials with randomly selected multi-view sets.
In Figs.~\ref{fig:results_case1_and_2} (a) and \ref{fig:results_case1_and_2} (d), which show the results with ModelNet40, we also draw the scores with the original MVCNN using Support Vector Machine (SVM) reported in~\cite{su15mvcnn}.
Interestingly, as we focus on the object classification task whereas Su~\etal~\cite{su15mvcnn} focused more on object retrieval task, we found that the baseline method with late view-pooling is slightly better in this case than the original MVCNN with the view-pooling layer in the middle.
The baseline method does especially well with ModelNet10 in case (i) (\fref{fig:results_case1_and_2} (b)), where it achieves the best performance among the methods. 
With ModelNet40 in case (i) (\fref{fig:results_case1_and_2} (a)), RotationNet achieved a comparable result with MVCNN when we used all the $12$ views as input.
In case (ii) (Figs.~\ref{fig:results_case1_and_2} (d) and (e)), where we consider full 3D rotation, RotationNet demonstrated superior performance to other methods. Only with three views, it showed comparable performance to that of MVCNN with a full set ($80$ views) of multi-view images.

Next, we investigate the performance of RotationNet with three different architectures: AlexNet~\cite{Krizhevsky_imagenetclassification}, VGG-M~\cite{Chatfield14}, and ResNet-50~\cite{he2015deep}.
\Tref{table:architectures_comparison} shows the classification accuracy on ModelNet40 and ModelNet10.
Because we deal with discrete viewpoints, we altered $11$ different camera system orientations (similarly to \cite{chen2003}) and calculated the mean and maximum accuracy of those trials.
Surprisingly, the performance difference among different architectures is marginal compared to the difference caused by different camera system orientations.
It indicates that the placement of viewpoints is the most important factor in multiview-based 3D object classification.
See \sref{sec:camerasystem} for more details.

\begin{table}[tb]
  \begin{center}
    \begin{tabular}{@{}lllll@{}}
      \toprule
      & \multicolumn{2}{l}{ModelNet40} & \multicolumn{2}{l}{ModelNet10} \\
      \midrule
          Archit. & Mean & Max & Mean & Max \\
      \midrule
      AlexNet & 93.70 $\pm$ 1.07 & 96.39 & 94.52 $\pm$ 1.01 & 97.58 \\
      VGG-M & 94.68 $\pm$ 1.16 & \textbf{97.37} & \textbf{94.82 $\pm$ 1.17} & \textbf{98.46} \\
      ResNet-50\hspace{-2mm} & \textbf{94.77 $\pm$ 1.10} & 96.92 & 94.80 $\pm$ 0.96 & 97.80 \\
      \bottomrule
    \end{tabular}
  \end{center}
  \vspace{-2mm}
  \caption{Comparison of classification accuracy (\%) with RotationNet based on different architectures.}
  \label{table:architectures_comparison}
\end{table}

Finally, we summarize the comparison of classification accuracy on ModelNet40 and ModelNet10 to existing 3D object classification methods in \Tref{table:classification}\footnote{We do not include the scores of ``VRN Ensemble''~\cite{brock2016generative} using ensembling technique because is written in ~\cite{brock2016generative} ``we suspect that this result is not general, and do not claim it with our main results.'' The reported scores are 95.54\% with ModelNet40 and 97.14\% with ModelNet10, which are both outperformed by RotationNet \textbf{with any architecture} (see \Tref{table:architectures_comparison}). }.
RotationNet (with VGG-M architecture) significantly outperformed existing methods with both the ModelNet40 and ModelNet10 datasets.
We reported the maximum accuracy among the aforementioned 11 rotation trials.
Note that the average accuracy of those trials on ModelNet40 was 94.68\%, which is still superior to the current state-of-the-art score 93.8\% reported by Wang~\etal~\cite{wang2017bmvc}.
Besides, Wang~\etal~\cite{wang2017bmvc} used additional feature modalities: surface normals and normalized depth values to improve the performance by $>1$\%.

\begin{figure*}[t]
  \begin{center}
    \begin{minipage}{.32\hsize}
        \begin{center}
          \begin{tabular}{@{}lll@{}}
            \toprule
            Algorithm	& class &	view \\
            \midrule
            MVCNN (softmax) & 86.08 & - \\
            Baseline & 88.73 & - \\
            Fine-grained, $T$=300 & 81.23 & 26.94 \\
            Fine-grained, $T$=4K  & 76.95 & 31.96 \\
            \textbf{RotationNet} & \textbf{89.31} & \textbf{33.59} \\
            \bottomrule
          \end{tabular}
        \end{center}
        \tblcaption{Accuracy of classification and viewpoint estimation (\%) in case (i) with RGBD.}
        \label{table:classification_viewestimation}
    \end{minipage}
    \hspace{0.5mm}
    \begin{minipage}{.32\hsize}
        \begin{center}
          \begin{tabular}{@{}lll@{}}
            \toprule
            Algorithm	& class &	view \\
            \midrule
            MVCNN (softmax) & 94.17 & - \\
            Baseline & 95 & - \\
            Fine-grained, $T$=800 & 92.76 & 56.72 \\
            Fine-grained, $T$=4K & 91.35 & 58.33 \\
            \textbf{RotationNet} & \textbf{98.33} & \textbf{85.83} \\
            \bottomrule
          \end{tabular}
        \end{center}
        \tblcaption{Accuracy of classification and viewpoint estimation (\%) in case (i) with MIRO.}
        \label{table:classification_viewestimation_2}
    \end{minipage}      
    \hspace{0.5mm}
    \begin{minipage}{.32\hsize}
        \begin{center}
          \begin{tabular}{@{}lll@{}}
            \toprule
            Algorithm	& class &	view \\
            \midrule
            MVCNN (softmax) & 95 & - \\
            Baseline & 95.83 & - \\
             Fine-grained, $T$=1.1K   & 94.21 & 70.63 \\
             Fine-grained, $T$=2.6K   & 93.54 & 72.38 \\
            \textbf{RotationNet} & \textbf{99.17} & \textbf{75.67} \\
            \bottomrule
          \end{tabular}
        \end{center}
        \tblcaption{Accuracy of classification and viewpoint estimation (\%) in case (ii) with MIRO.}
        \label{table:classification_viewestimation_3}
    \end{minipage}      
  \end{center}
\end{figure*}

\begin{table}[tb]
  \begin{center}
    \begin{tabular}{@{}lll@{}}
      \toprule
          Algorithm	& {\small ModelNet40} &	{\small ModelNet10} \\
      \midrule
      \textbf{RotationNet} & \textbf{97.37} & \textbf{98.46} \\
      Dominant Set Clustering~\cite{wang2017bmvc} & 93.8 & - \\
      Kd-Networks~\cite{klokov2017} & 91.8 & 94.0 \\
      MVCNN-MultiRes~\cite{Qi_2016_CVPR} & 91.4 & - \\
      ORION~\cite{sedaghat2016orientation} & - & 93.80 \\
      VRN~\cite{brock2016generative} & 91.33 & 93.61 \\
      FusionNet~\cite{hegde2016fusionnet} & 90.80 & 93.11 \\
      Pairwise~\cite{johns2016pairwise} & 90.70 & 92.80 \\
      PANORAMA-NN~\cite{sfikas2017} & 90.7 & 91.1 \\
      MVCNN~\cite{su15mvcnn} & 90.10 & - \\
      Set-convolution~\cite{ravanbakhsh2016deep} & 90 & - \\
      FPNN~\cite{li_fpnn2016} & 88.4 & - \\
      Multiple Depth Maps~\cite{zanuttigh2017} & 87.8 & 91.5 \\
      LightNet~\cite{zhi2017} & 86.90 & 93.39 \\
      PointNet~\cite{qi2016pointnet} & 86.2 & - \\
      Geometry Image~\cite{sinha2016deep} & 83.9 & 88.4 \\
      3D-GAN~\cite{Jiajun2016} & 83.30 & 91.00 \\
      ECC~\cite{simonovsky2017} & 83.2 & - \\
      GIFT~\cite{bai2016gift} & 83.10 & 92.35 \\
      VoxNet~\cite{maturana2015voxnet} & 83 & 92 \\
      Beam Search~\cite{xu2016beam} & 81.26 & 88 \\
      DeepPano~\cite{shi2015deeppano} & 77.63 & 85.45 \\
      3DShapeNets~\cite{wu20153d} & 77 & 83.50 \\
      PointNet~\cite{garcia2016pointnet} & - & 77.6 \\
      \bottomrule
    \end{tabular}
  \end{center}
  \vspace{-2mm}
  \caption{Comparison of classification accuracy (\%). RotationNet achieved the {\it state-of-the-art} performance both with ModelNet40 and ModelNet10.}
  \label{table:classification}
\end{table}

\subsection{Experiment on a real image benchmark dataset}
\label{subsec:1DOF}
Next, we describe the experimental results on a benchmark RGBD dataset published in~\cite{lai2011large}, which consists of real images of objects on a one-dimensional rotation table.
This dataset contains $300$ object instances in $51$ categories.
Although it contains depth images and 3D point clouds, we used only RGB images in our experiment.
We applied the upright orientation assumption (case (i)) in this experiment, because the bottom faces of objects on the turning table were not recorded.
We picked out $12$ images of each object instance with the closest rotation angles to $\{0^\circ, 30^\circ, \cdots, 330^\circ\}$.
In the training phase, objects are self-aligned (in an unsupervised manner) and the viewpoint variables for images are determined. 
To predict the pose of a test object instance, we predict the discrete viewpoint that each test image is observed, 
and then refer the most frequent pose value among those attached to the training samples predicted to be observed from the same viewpoint.



\Tref{table:classification_viewestimation} summarizes the classification and viewpoint estimation accuracies.
The baseline method and MVCNN are not able to estimate viewpoints because they are essentially viewpoint invariant.
As another baseline approach to compare, we learned a CNN with AlexNet architecture that outputs $612$ $(= 51 \times 12)$ scores to distinguish both viewpoints and categories, which we call ``Fine-grained.''
Here, $T$ denotes the number of iterations that the CNN parameters are updated in the training phase.
As shown in \Tref{table:classification_viewestimation}, the classification accuracy with ``Fine-grained'' decreases while its viewpoint estimation accuracy improves as the iteration grows.
We consider this is because the ``Fine-grained'' classifiers become more and more sensitive to intra-class appearance variation through training, which affects the categorization accuracy.
In contrast, RotationNet demonstrated the best performance in both object classification and viewpoint estimation, although the ground-truth poses are not given to RotationNet during the training.

\Tref{table:pose_estimation} shows the object instance/category recognition as well as pose estimation accuracy comparison to existing methods.
RotationNet with a single image input performs comparable to Elhoseiny~\etal~\cite{elhoseiny2016comparative}.
Interestingly, when we estimate object instance/category and pose using $12$ views altogether, both accuracies are remarkably improved.

\begin{table}[tb]
  \small
  \begin{center}
    \begin{tabular}{@{}lrrr@{}}
      \toprule
      \hspace{-4mm}&\hspace{-4mm} Instance (\%) \hspace{-1mm} & \hspace{-1mm} Category (\%) \hspace{-1mm} & \hspace{-1mm} Avg. Pose (\%) \\
      \midrule
          Lai~\etal~\cite{lai_aaai11} & 78.40 & 94.30 & 53.50 \\
          Zhang~\etal~\cite{zhang2013joint} & 74.79 & 93.10 & 61.57 \\
          Bakry~\etal~\cite{bakry2014untangling} & 80.10 & 94.84 & 76.63 \\
          Elhoseiny~\etal~\cite{elhoseiny2016comparative} & - & \underline{97.14} & \underline{79.30} \\
          Ours - single view & \underline{90.44} & 96.55 & 78.67 \\
          Ours - 12 views & \textbf{97.45} & \textbf{99.51} & \textbf{81.17} \\
          \bottomrule
    \end{tabular}
  \end{center}
  \vspace{-2mm}
  \caption{Comparison on object instance/category recognition and pose estimation on RGBD dataset.}
  \label{table:pose_estimation}
  \normalsize
\end{table}



\subsection{Experiment on a 3D rotated real image dataset}
\label{subsec:2DOF}

\begin{figure*}[t]
  \begin{center}

    \includegraphics[width=.49\linewidth]{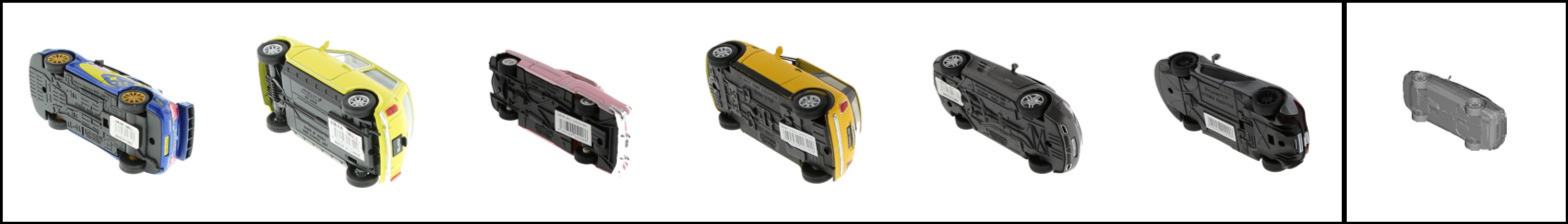}
    \includegraphics[width=.49\linewidth]{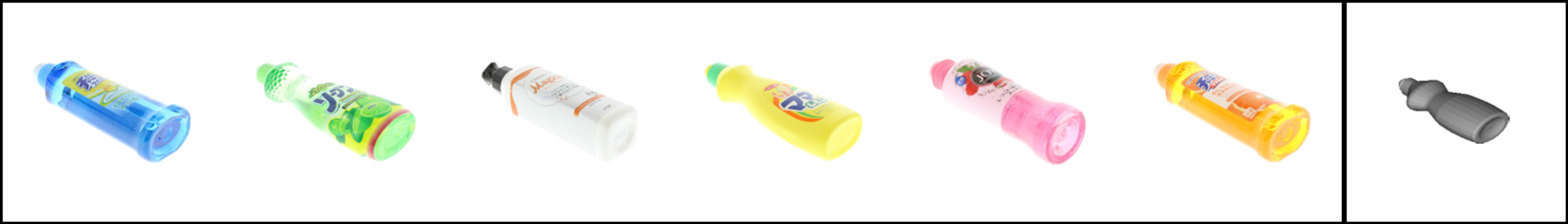}
    \includegraphics[width=.49\linewidth]{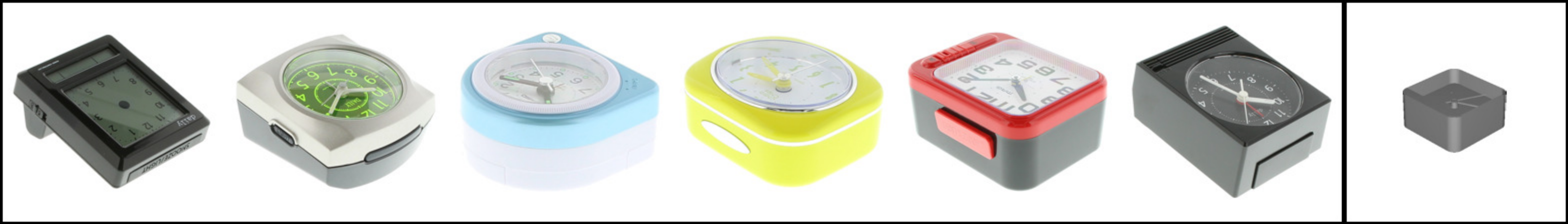}
    \includegraphics[width=.49\linewidth]{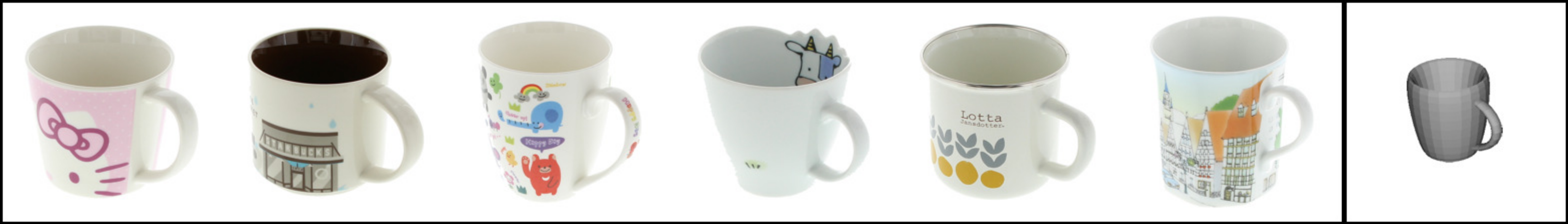}
    \includegraphics[width=.49\linewidth]{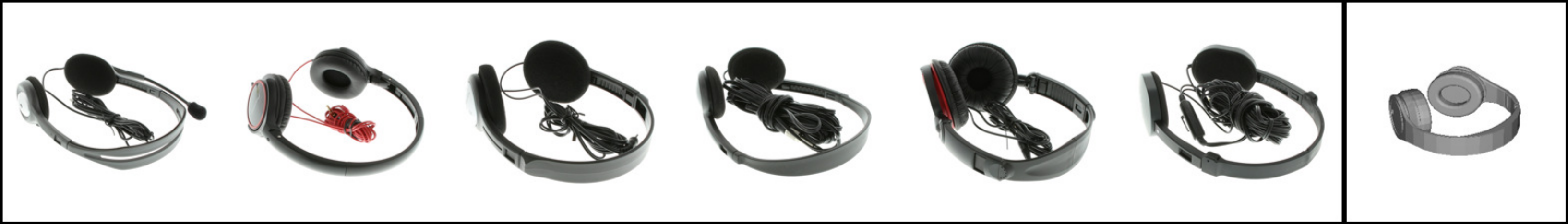}
    \includegraphics[width=.49\linewidth]{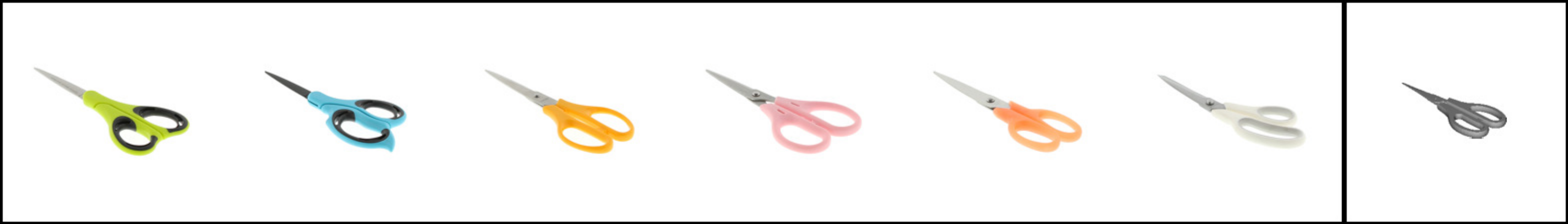}
    \includegraphics[width=.49\linewidth]{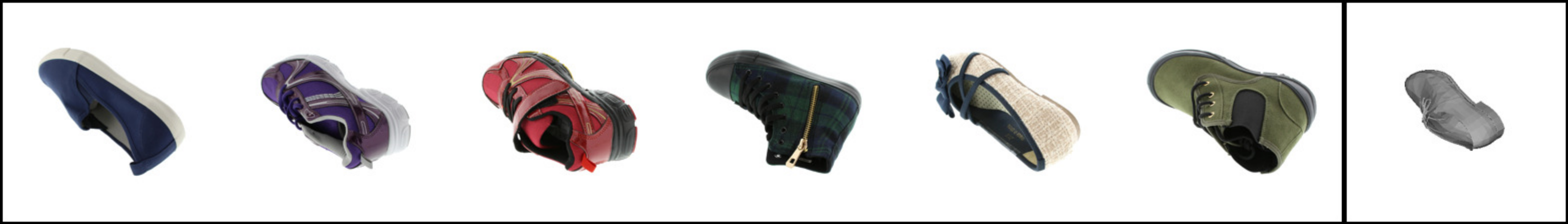}
    \includegraphics[width=.49\linewidth]{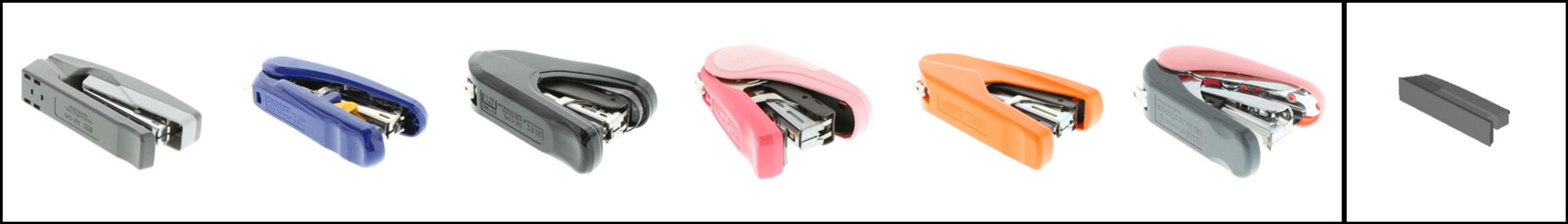}
    \includegraphics[width=.49\linewidth]{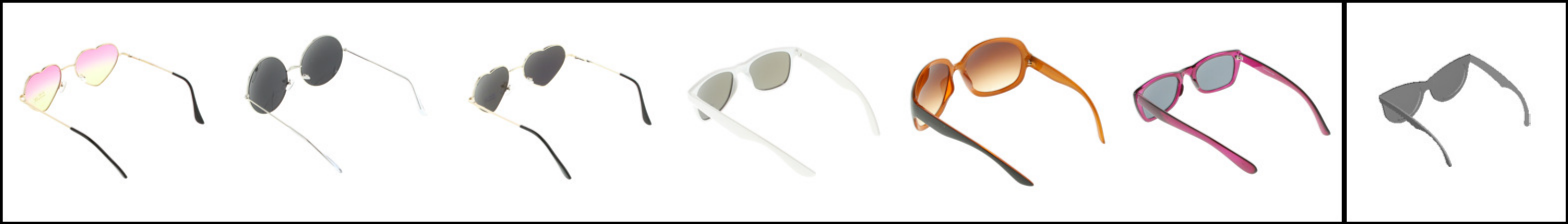}
    \includegraphics[width=.49\linewidth]{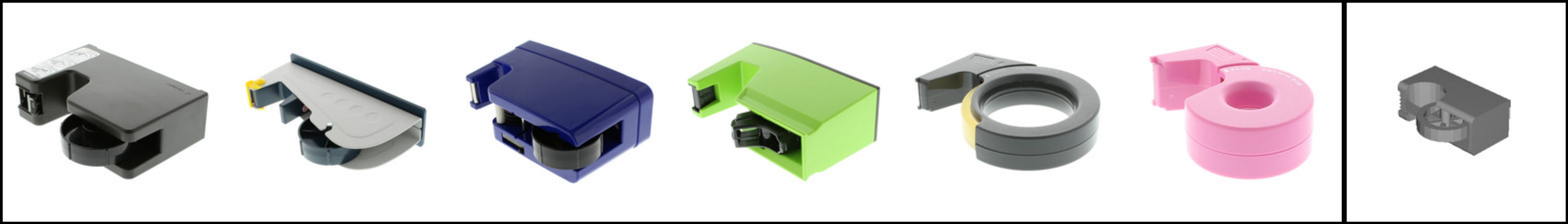}
    \includegraphics[width=.49\linewidth]{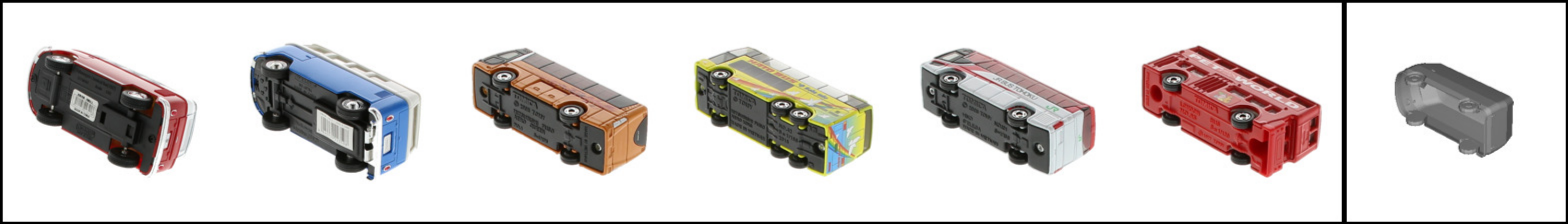}
    \includegraphics[width=.49\linewidth]{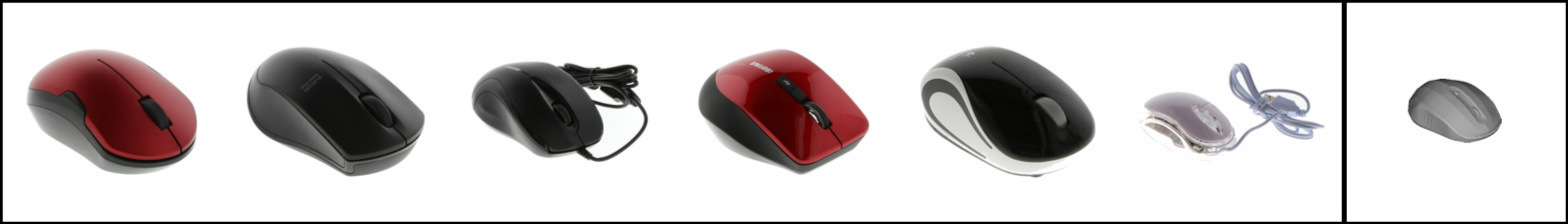}

  \end{center}
  \vspace{-4mm}
  \caption{Object instances self-aligned as a result of training RotationNet. The last instance in each category is a 3D CAD model. }
  \label{fig:alignment_examples}
\end{figure*}

\begin{figure*}[t]
  \begin{center}
    \includegraphics[width=\linewidth]{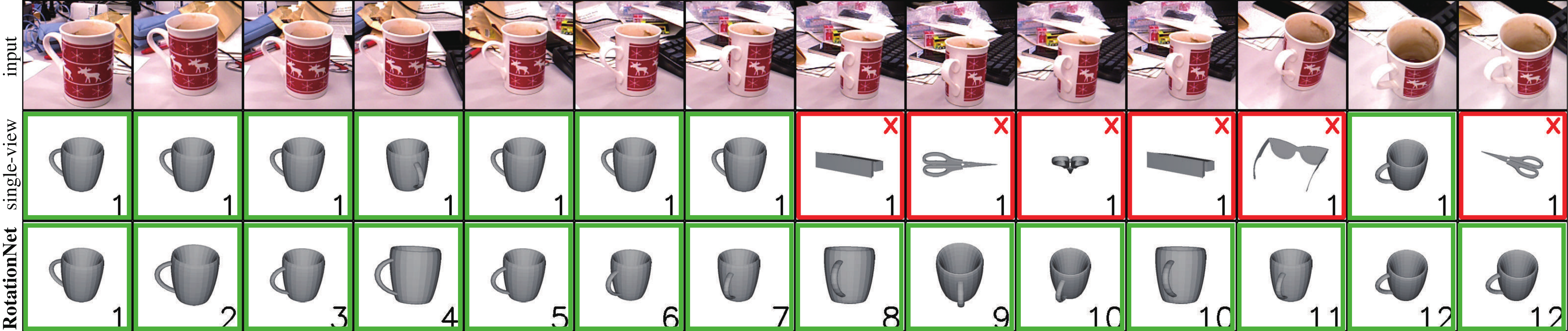}
    \vspace{2mm}
    \includegraphics[width=\linewidth]{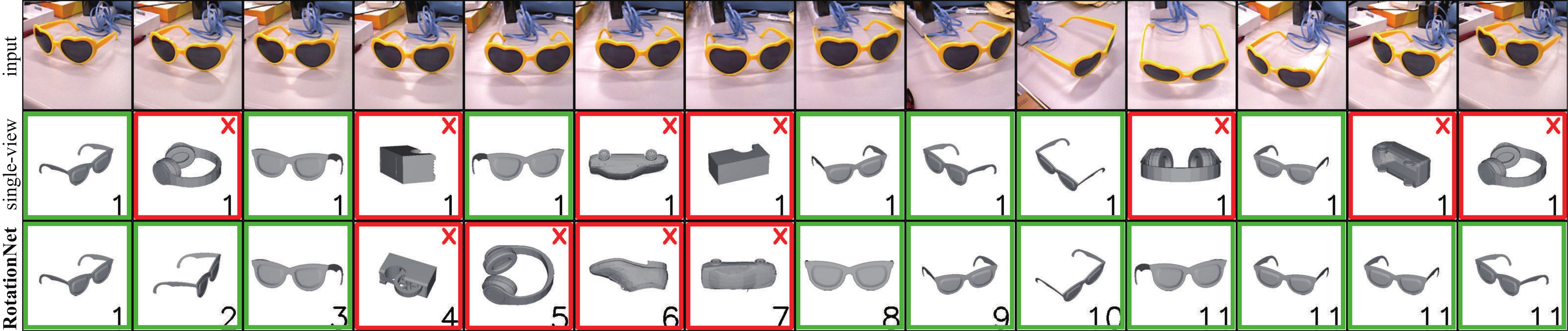}
  \end{center}
  \vspace{-4mm}
  \caption{Exemplar objects recognized using a USB camera. The second and fifth rows show 3D models in the estimated category and pose from a single view, whereas the third and sixth rows show those estimated using multiple views. The number in each image indicates the number of views used for predictions. Failure cases are shown in red boxes. See the video in the supplementary material for more qualitative results. The video also contains the real-time demonstration with the Microsoft HoloLens device.
}
  \label{fig:realworld_examples}
\end{figure*}

We describe the experimental results on our new dataset ``Multi-view Images of Rotated Objects (MIRO)'' in this section.
Exemplar images in MIRO dataset are shown in \sref{sec:MIROinstances}.
We used Ortery's 3D MFP studio\footnote{https://www.ortery.com/photography-equipment/3d-modeling/} to capture multi-view images of objects with 3D rotations.
The RGBD benchmark dataset~\cite{lai2011large} has two issues for training multi-view based CNNs: insufficient number of object instances per category (which is a minimum of two for training) and inconsistent cases to the upright orientation assumption.
There are several cases where the upright orientation assumption is actually invalid; the attitudes of object instances against the rotation axis are inconsistent in some object categories.
Also, this dataset does not include the bottom faces of objects on the turning table.
Our MIRO dataset includes $10$ object instances per object category. It consists of $120$ object instances in $12$ categories in total.
We captured each object instance with $M_e = 10$ levels of elevation angles and $16$ levels of azimuth angles to obtain $160$ images.
For our experiments, we used $16$ images ($\theta=22.5^\circ$) with $0^\circ$ elevation of an object instance in case (i).
We carefully captured all the object instances in each category to have the same upright direction in order to evaluate performance in the case (i).
For case (ii), we used $20$ images observed from the $20$ vertices of a dodecahedron encompassing an object.

Figures~\ref{fig:results_case1_and_2} (c) and \ref{fig:results_case1_and_2} (f) show the object classification accuracy versus the number of views used for the prediction in case (i) and case (ii), respectively.
In both cases, RotationNet clearly outperforms both MVCNN and the baseline method when the number of views is larger than $2$.
We also tested the ``Fine-grained'' method that outputs $(192=12\times16)$ scores in case (i) and $(240=12\times20)$ scores in case (ii) to distinguish both viewpoints and categories, and the overall results are summarized in Tables~\ref{table:classification_viewestimation_2} and \ref{table:classification_viewestimation_3}.
Similar to the results with an RGBD dataset described above, there is a trade-off between object classification and viewpoint estimation accuracies in the ``Fine-grained'' approach.
RotationNet achieved the best performance in both object classification and viewpoint estimation, which demonstrates the strength of the proposed approach.
%



Finally, we demonstrate the performance of RotationNet for real-world applications.
For training, we used our MIRO dataset with the viewpoint setup case (\rnum{3}), where all the outputs for images with $10$ levels of elevation angles are concatenated,
which enables RotationNet to distinguish $160$ viewpoints.
We added rendered images of a single 3D CAD model (whose upright orientation is manually assigned) to each object class, which were trained together with MIRO dataset.
Then we obtained successful alignments between a CAD model and real images for all the 12 object classes (\fref{fig:alignment_examples}).
\Fref{fig:realworld_examples} shows exemplar objects recognized using a USB camera.
We estimated relative camera poses by LSD-SLAM~\cite{engel2014lsd} to integrate predictions from multiple views in sequence.
The results obtained using multiple views (shown in the third and sixth rows) are consistently more accurate than those using a single view (shown in the second and fifth rows).
It is worth noting that not only object classification but also pose estimation performance is improved by using multiple views.

\section{Discussion}

We proposed RotationNet, which jointly estimates object category and viewpoint from each single-view image and aggregates the object class predictions obtained from a partial set of multi-view images.
In our method, object instances are automatically aligned in an unsupervised manner with both inter-class and intra-class structures based on their appearance during the training.
In the experiment using 3D object benchmark datasets ModelNet40 and ModelNet10, RotationNet significantly outperformed the state-of-the-art methods based on voxels, point clouds, and multi-view images.
RotationNet is also able to achieve comparable performance to MVCNN~\cite{su15mvcnn} with $80$ different multi-view images using only a couple of view images, which is important for real-world applications.
Another contribution is that we developed a publicly available new dataset named MIRO. 
Using this dataset and RGBD object benchmark dataset~\cite{lai2011large}, we showed that RotationNet even outperformed supervised learning based approaches in a pose estimation task.
We consider that our pose estimation performance benefits from view-specific appearance information shared across classes due to the inter-class self-alignment.

Similar to MVCNN~\cite{su15mvcnn} and any other 3D object classification method that considers discrete variance of rotation, RotationNet has the limitation that each image should be observed from one of the pre-defined viewpoints. 
The discrete pose estimation by RotationNet, however, demonstrated superior performance to existing methods on the RGBD object benchmark dataset.
It can be further improved by introducing a fine pose alignment post-process using \eg iterative closest point (ICP) algorithm.
Another potential avenue to look into is the automatic selection of the best camera system orientations, since it has an effect on object classification accuracy.

\section*{Acknowledgment}
This project is supported by the New Energy and Industrial Technology Development Organization (NEDO).
The authors would like to thank Hiroki Matsuno for his support in developing the HoloLens application.

{\small
\bibliographystyle{ieee}
\bibliography{rotationnet}

\begin{thebibliography}{10}\itemsep=-1pt

\bibitem{bai2016gift}
S.~Bai, X.~Bai, Z.~Zhou, Z.~Zhang, and L.~J. Latecki.
\newblock Gift: A real-time and scalable 3d shape search engine.
\newblock In {\em {Proceedings of IEEE Conference on Computer Vision and
  Pattern Recognition (CVPR)}}, 2016.

\bibitem{bakry2014untangling}
A.~Bakry and A.~Elgammal.
\newblock Untangling object-view manifold for multiview recognition and pose
  estimation.
\newblock In {\em {Proceedings of European Conference on Computer Vision
  (ECCV)}}, 2014.

\bibitem{bo2013unsupervised}
L.~Bo, X.~Ren, and D.~Fox.
\newblock Unsupervised feature learning for rgb-d based object recognition.
\newblock In {\em Proceedings of International Symposium on Experimental
  Robotics (ISER)}, 2013.

\bibitem{borotschnig2000appearance}
H.~Borotschnig, L.~Paletta, M.~Prantl, and A.~Pinz.
\newblock Appearance-based active object recognition.
\newblock {\em Image and Vision Computing}, 18(9), 2000.

\bibitem{brock2016generative}
A.~Brock, T.~Lim, J.~Ritchie, and N.~Weston.
\newblock Generative and discriminative voxel modeling with convolutional
  neural networks.
\newblock In {\em Proceedings of NIPS Workshop on 3D Deep Learning}, 2017.

\bibitem{Chatfield14}
K.~Chatfield, K.~Simonyan, A.~Vedaldi, and A.~Zisserman.
\newblock Return of the devil in the details: Delving deep into convolutional
  nets.
\newblock In {\em {Proceedings of British Machine Vision Conference (BMVC)}},
  2014.

\bibitem{chen2014inferring}
C.-Y. Chen and K.~Grauman.
\newblock Inferring unseen views of people.
\newblock In {\em {Proceedings of IEEE Conference on Computer Vision and
  Pattern Recognition (CVPR)}}, 2014.

\bibitem{chen2003}
D.-Y. Chen, X.-P. Tian, Y.-T. Shen, and M.~Ouhyoung.
\newblock On visual similarity based 3{D} model retrieval.
\newblock {\em Computer Graphics Forum}, 22(3), 2003.

\bibitem{elhoseiny2016comparative}
M.~Elhoseiny, T.~El-Gaaly, A.~Bakry, and A.~Elgammal.
\newblock A comparative analysis and study of multiview cnn models for joint
  object categorization and pose estimation.
\newblock In {\em {Proceedings of International Conference on Machine Learning
  (ICML)}}, 2016.

\bibitem{engel2014lsd}
J.~Engel, T.~Sch{\"o}ps, and D.~Cremers.
\newblock Lsd-slam: Large-scale direct monocular slam.
\newblock In {\em {Proceedings of European Conference on Computer Vision
  (ECCV)}}, 2014.

\bibitem{garcia2016pointnet}
A.~Garcia-Garcia, F.~Gomez-Donoso, J.~Garcia-Rodriguez, S.~Orts-Escolano,
  M.~Cazorla, and J.~Azorin-Lopez.
\newblock Pointnet: A 3d convolutional neural network for real-time object
  class recognition.
\newblock In {\em Proceedings of IEEE International Joint Conference on Neural
  Networks (IJCNN)}, 2016.

\bibitem{he2015deep}
K.~He, X.~Zhang, S.~Ren, and J.~Sun.
\newblock Deep residual learning for image recognition.
\newblock In {\em {Proceedings of IEEE Conference on Computer Vision and
  Pattern Recognition (CVPR)}}, 2016.

\bibitem{hegde2016fusionnet}
V.~Hegde and R.~Zadeh.
\newblock Fusionnet: 3d object classification using multiple data
  representations.
\newblock {\em arXiv preprint arXiv:1607.05695}, 2016.

\bibitem{johns2016pairwise}
E.~Johns, S.~Leutenegger, and A.~J. Davison.
\newblock Pairwise decomposition of image sequences for active multi-view
  recognition.
\newblock In {\em {Proceedings of IEEE Conference on Computer Vision and
  Pattern Recognition (CVPR)}}, 2016.

\bibitem{klokov2017}
R.~Klokov and V.~Lempitsky.
\newblock Escape from cells: Deep kd-networks for the recognition of 3d point
  cloud models.
\newblock In {\em {Proceedings of International Conference on Computer Vision
  (ICCV)}}, 2017.

\bibitem{Krizhevsky_imagenetclassification}
A.~Krizhevsky, I.~Sutskever, and G.~E. Hinton.
\newblock Imagenet classification with deep convolutional neural networks.
\newblock In {\em {Proceedings of Advances in Neural Information Processing
  Systems (NIPS)}}, 2012.

\bibitem{Kuznetsova2016}
A.~Kuznetsova, S.~J. Hwang, B.~Rosenhahn, and L.~Sigal.
\newblock Exploiting view-specific appearance similarities across classes for
  zero-shot pose prediction: A metric learning approach.
\newblock In {\em {Proceedings of AAAI Conference on Artificial Intelligence}},
  2016.

\bibitem{lai2011large}
K.~Lai, L.~Bo, X.~Ren, and D.~Fox.
\newblock A large-scale hierarchical multi-view rgb-d object dataset.
\newblock In {\em {Proceedings of IEEE International Conference on Robotics and
  Automation (ICRA)}}. IEEE, 2011.

\bibitem{lai_aaai11}
K.~Lai, L.~Bo, X.~Ren, and D.~Fox.
\newblock A scalable tree-based approach for joint object and pose recognition.
\newblock In {\em {Proceedings of AAAI Conference on Artificial Intelligence}},
  2011.

\bibitem{li_fpnn2016}
Y.~Li, S.~Pirk, H.~Su, C.~R. Qi, , and L.~J. Guibas.
\newblock Fpnn: Field probing neural networks for 3d data.
\newblock In {\em {Proceedings of Advances in Neural Information Processing
  Systems (NIPS)}}, 2016.

\bibitem{maturana2015voxnet}
D.~Maturana and S.~Scherer.
\newblock Voxnet: A 3d convolutional neural network for real-time object
  recognition.
\newblock In {\em {Proceedings of IEEE/RSJ International Conference on
  Intelligent Robots and Systems (IROS)}}, 2015.

\bibitem{Novotny_2017_ICCV}
D.~Novotny, D.~Larlus, and A.~Vedaldi.
\newblock Learning 3d object categories by looking around them.
\newblock In {\em {Proceedings of International Conference on Computer Vision
  (ICCV)}}, 2017.

\bibitem{paletta2000active}
L.~Paletta and A.~Pinz.
\newblock Active object recognition by view integration and reinforcement
  learning.
\newblock {\em Robotics and Autonomous Systems}, 31(1), 2000.

\bibitem{qi2016pointnet}
C.~R. Qi, H.~Su, K.~Mo, and L.~J. Guibas.
\newblock Pointnet: Deep learning on point sets for 3{D} classification and
  segmentation.
\newblock In {\em {Proceedings of IEEE Conference on Computer Vision and
  Pattern Recognition (CVPR)}}, 2017.

\bibitem{Qi_2016_CVPR}
C.~R. Qi, H.~Su, M.~Niessner, A.~Dai, M.~Yan, and L.~J. Guibas.
\newblock Volumetric and multi-view {CNN}s for object classification on 3{D}
  data.
\newblock In {\em {Proceedings of IEEE Conference on Computer Vision and
  Pattern Recognition (CVPR)}}, 2016.

\bibitem{ravanbakhsh2016deep}
S.~Ravanbakhsh, J.~Schneider, and B.~Poczos.
\newblock Deep learning with sets and point clouds.
\newblock {\em arXiv preprint arXiv:1611.04500}, 2016.

\bibitem{ILSVRC15}
O.~Russakovsky, J.~Deng, H.~Su, J.~Krause, S.~Satheesh, S.~Ma, Z.~Huang,
  A.~Karpathy, A.~Khosla, M.~Bernstein, A.~C. Berg, and L.~Fei-Fei.
\newblock {ImageNet Large Scale Visual Recognition Challenge}.
\newblock {\em {International Journal of Computer Vision}}, 115(3), 2015.

\bibitem{savarese20073d}
S.~Savarese and L.~Fei-Fei.
\newblock 3{D} generic object categorization, localization and pose estimation.
\newblock In {\em {Proceedings of International Conference on Computer Vision
  (ICCV)}}, 2007.

\bibitem{sedaghat2016orientation}
N.~Sedaghat, M.~Zolfaghari, and T.~Brox.
\newblock Orientation-boosted voxel nets for 3{D} object recognition.
\newblock In {\em {Proceedings of British Machine Vision Conference (BMVC)}},
  2017.

\bibitem{sfikas2017}
K.~Sfikas, T.~Theoharis, and I.~Pratikakis.
\newblock Exploiting the panorama representation for convolutional neural
  network classification and retrieval.
\newblock In {\em Proceedings of Eurographics Workshop on 3D Object Retrieval
  (3DOR)}, 2017.

\bibitem{shi2015deeppano}
B.~Shi, S.~Bai, Z.~Zhou, and X.~Bai.
\newblock Deeppano: Deep panoramic representation for 3-d shape recognition.
\newblock {\em IEEE Signal Processing Letters}, 22(12), 2015.

\bibitem{simonovsky2017}
M.~Simonovsky and N.~Komodakis.
\newblock Dynamic edge-conditioned filters in convolutional neural networks on
  graphs.
\newblock In {\em {Proceedings of IEEE Conference on Computer Vision and
  Pattern Recognition (CVPR)}}, 2017.

\bibitem{sinha2016deep}
A.~Sinha, J.~Bai, and K.~Ramani.
\newblock Deep learning 3{D} shape surfaces using geometry images.
\newblock In {\em {Proceedings of European Conference on Computer Vision
  (ECCV)}}, 2016.

\bibitem{su15mvcnn}
H.~Su, S.~Maji, E.~Kalogerakis, and E.~G. Learned-Miller.
\newblock Multi-view convolutional neural networks for 3{D} shape recognition.
\newblock In {\em {Proceedings of International Conference on Computer Vision
  (ICCV)}}, 2015.

\bibitem{Su_2015_ICCV}
H.~Su, C.~R. Qi, Y.~Li, and L.~J. Guibas.
\newblock Render for cnn: Viewpoint estimation in images using cnns trained
  with rendered 3{D} model views.
\newblock In {\em {Proceedings of International Conference on Computer Vision
  (ICCV)}}, 2015.

\bibitem{su20153d}
H.~Su, F.~Wang, E.~Yi, and L.~J. Guibas.
\newblock 3{D}-assisted feature synthesis for novel views of an object.
\newblock In {\em {Proceedings of International Conference on Computer Vision
  (ICCV)}}, 2015.

\bibitem{wang2017bmvc}
C.~Wang, M.~Pelillo, and K.~Siddiqi.
\newblock Dominant set clustering and pooling for multi-view 3d object
  recognition.
\newblock In {\em {Proceedings of British Machine Vision Conference (BMVC)}},
  2017.

\bibitem{Jiajun2016}
J.~Wu, C.~Zhang, T.~Xue, W.~T. Freeman, and J.~B. Tenenbaum.
\newblock Learning a probabilistic latent space of object shapes via 3{D}
  generative-adversarial modeling.
\newblock In {\em {Proceedings of Advances in Neural Information Processing
  Systems (NIPS)}}, 2016.

\bibitem{wu20153d}
Z.~Wu, S.~Song, A.~Khosla, F.~Yu, L.~Zhang, X.~Tang, and J.~Xiao.
\newblock 3d shapenets: A deep representation for volumetric shapes.
\newblock In {\em {Proceedings of IEEE Conference on Computer Vision and
  Pattern Recognition (CVPR)}}, 2015.

\bibitem{xu2016beam}
X.~Xu and S.~Todorovic.
\newblock Beam search for learning a deep convolutional neural network of 3d
  shapes.
\newblock In {\em {Proceedings of International Conference on Pattern
  Recognition (ICPR)}}, 2016.

\bibitem{zanuttigh2017}
P.~Zanuttigh and L.~Minto.
\newblock Deep learning for 3d shape classification from multiple depth maps.
\newblock In {\em {Proceedings of IEEE International Conference on Image
  Processing (ICIP)}}, 2017.

\bibitem{zhang2013joint}
H.~Zhang, T.~El-Gaaly, A.~M. Elgammal, and Z.~Jiang.
\newblock Joint object and pose recognition using homeomorphic manifold
  analysis.
\newblock In {\em {Proceedings of AAAI Conference on Artificial Intelligence}},
  volume~2, 2013.

\bibitem{zhi2017}
S.~Zhi, Y.~Liu, X.~Li, and Y.~Guo.
\newblock Lightnet: A lightweight 3{D} convolutional neural network for
  real-time 3{D} object recognition.
\newblock In {\em Proceedings of Eurographics Workshop on 3D Object Retrieval
  (3DOR)}, 2017.

\bibitem{zhou2017unsupervised}
T.~Zhou, M.~Brown, N.~Snavely, and D.~G. Lowe.
\newblock Unsupervised learning of depth and ego-motion from video.
\newblock In {\em {Proceedings of IEEE Conference on Computer Vision and
  Pattern Recognition (CVPR)}}, 2017.

\bibitem{NIPS2014_5546}
Z.~Zhu, P.~Luo, X.~Wang, and X.~Tang.
\newblock Multi-view perceptron: a deep model for learning face identity and
  view representations.
\newblock In {\em {Proceedings of Advances in Neural Information Processing
  Systems (NIPS)}}, 2014.

\end{thebibliography}
}

\onecolumn
\appendix

\section{Qualitative evaluation of self-alignment during the training of RotationNet}
\label{sec:selfalign}
\begin{figure}[h!]
  \begin{center}
    \includegraphics[width=\linewidth]{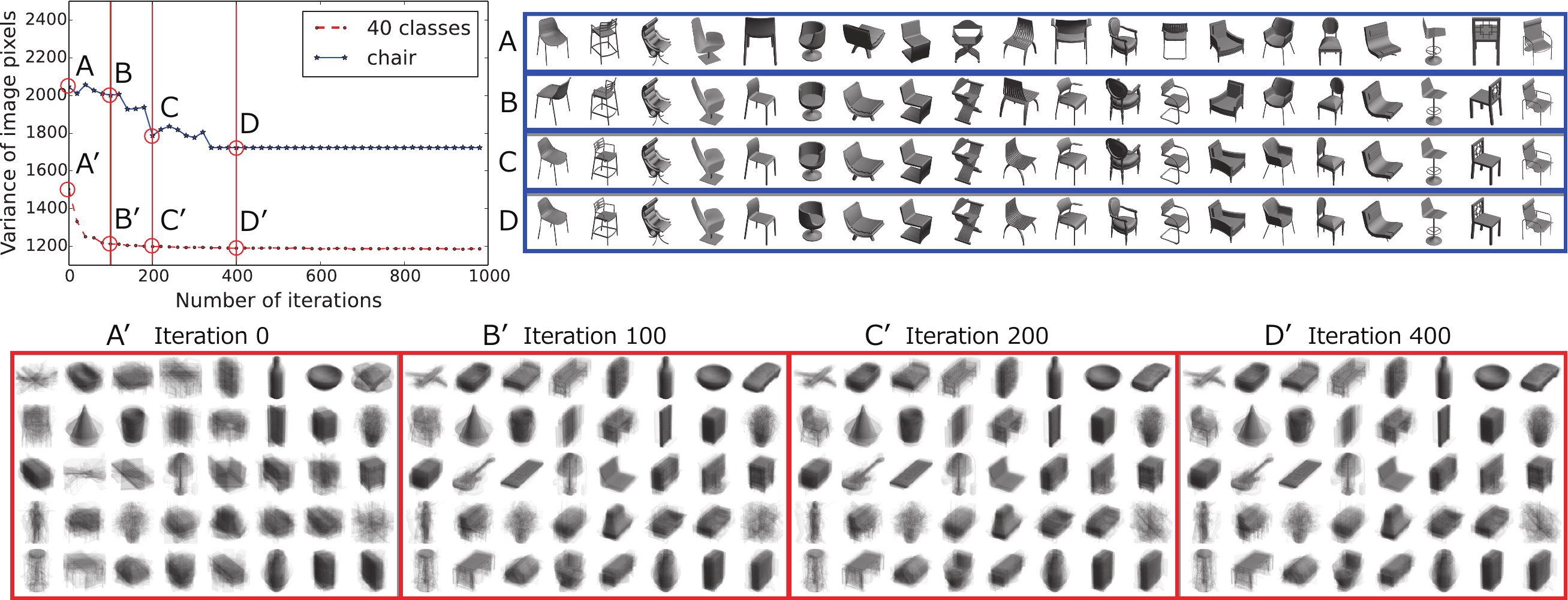}
  \end{center}
  \caption{Variance change of the average images generated by concatenating multi-view images (case (i)) in order of their predicted viewpoint variables. Average images of 40 classes and images of the ``chair'' class with the same predicted viewpoint variable in iterations 0, 100, 200, and 400 are shown in red boxes and blue boxes, respectively. }
  \label{fig:img_var_40v1_graph}
\end{figure}

\begin{figure}[t]
  \begin{center}
    \includegraphics[width=\linewidth]{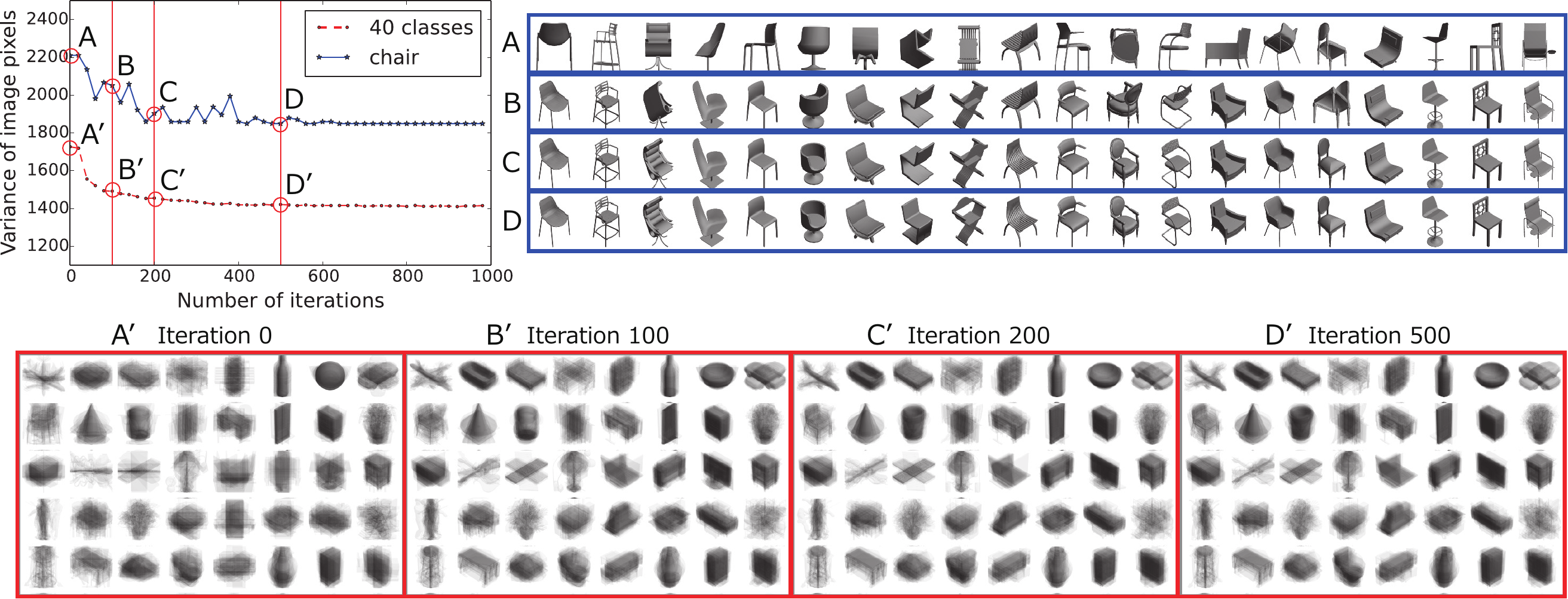}
  \end{center}
  \caption{Variance change of the average images generated by concatenating multi-view images (case (ii)) in order of their predicted viewpoint variables. Average images of 40 classes and images of the ``chair'' class with the same predicted viewpoint variable in iterations 0, 100, 200, and 500 are shown in red boxes and blue boxes, respectively. }
  \label{fig:img_var_40v2_graph}
\end{figure}

\vspace{-2mm}
Figures~\ref{fig:img_var_40v1_graph} and \ref{fig:img_var_40v2_graph} show the state transition of the inter- and intra-class object pose alignment that is automatically achieved during the training of RotationNet with ModelNet40, which depict the variation of the average images generated by concatenating multi-view images in order of their predicted viewpoint variables.
The figures correspond to cases (i): with upright orientation and (ii): w/o upright orientation, respectively. We can see that the variance of average images decreases together with the variance of object poses.
The red dotted lines show the mean variance of average images of all the $40$ classes, whereas the blue lines show the variance of average images of the ``chair'' class.

The images of test object instances in the ``chair'' class with the same predicted viewpoint variable are shown in the right of the figures.
In both cases (i) and (ii), where the latter case is more interesting because of its difficulty, the chairs with initial random poses gradually get aligned in the same direction after several hundreds of training iterations.
Moreover, the average images of all the 40 classes with the same predicted viewpoint variable shown in red boxes indicate that not only the intra-class alignment but also the inter-class alignment is achieved.
The alignment is less obvious in the red boxes of \fref{fig:img_var_40v2_graph}; however, it is confirmed that this does not harm the object classification accuracy.


\section{Influence of camera system orientation}
\label{sec:camerasystem}

\begin{figure}[h!]
  \begin{center}
    \includegraphics[width=\linewidth]{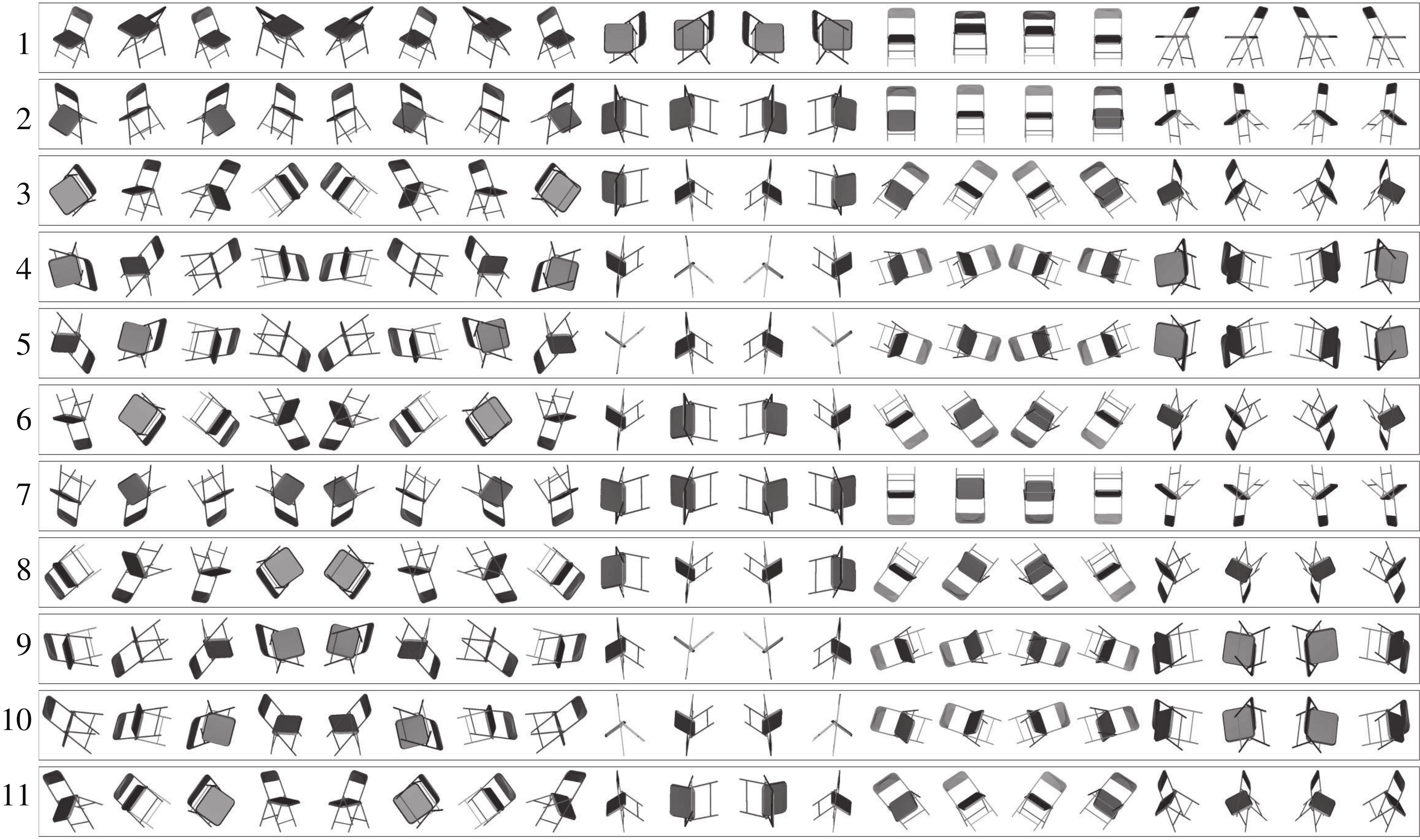}
  \end{center}
  \caption{Exemplar multi-view images of a chair captured in case (ii) with $11$ different camera system orientations. Numbers in the left indicate the camera system orientation IDs. }
  \label{fig:camera_orientation_illustration}
\end{figure}

As shown in Section 4.1 in the main manuscript, we tested the performance of RotationNet with $11$ different camera system orientations.
\Fref{fig:camera_orientation_illustration} shows exemplar multi-view images of a chair in ModelNet40 dataset captured in case (ii) with the $11$ camera system orientations.
Although ``aligned'' ModelNet40 dataset has been recently released, we used the original ``unaligned'' ModelNet40 dataset in our work.
Camera system orientations are first rotated by $36^\circ$ about the $x$-axis, and then rotated by $t \times 36^\circ$ $(t=0,\dots,9)$ about the $y$-axis.
In this way, different camera system orientations can capture different object profiles.
\Tref{table:camera_system_comparison} shows the comparison of classification accuracy (\%) on ModelNet40 and ModelNet10 with the different camera system orientations.
We altered the base architecture of RotationNet as AlexNet, VGG-M, and ResNet-50.
The best scores with each architecture among the orientations are shown in bold.
As shown here, the best camera system orientation is consistent across different architectures: the second one for ModelNet10 and the fourth one for ModelNet40.
It indicates that multi-view object classification is greatly improved by observing appropriate aspects of objects.
In addition, \Tref{table:camera_system_comparison} shows that the best performance on the validation set (which we extracted from the training split of ModelNet40) was achieved with the same camera system orientation as the test set.
Therefore, it is possible to obtain the best RotationNet model by selecting the one that best classifies a validation set among different camera system orientations.

\begin{table}[tb]
  \footnotesize
  \begin{center}
    \begin{tabular}{@{}llllllllllllll@{}}
      \toprule
      &  & \multicolumn{11}{c}{Camera system orientation ID} &  \\
      \cmidrule{3-13}
      Dataset & Archit. & 1 & 2 & 3 & 4 & 5 & 6 & 7 & 8 & 9 & 10 & 11 & Mean \\
      \midrule
      ModelNet40 - val & AlexNet & 93.03 & 94.33 & 95.14 & \textbf{95.54} & 92.63 & 92.46 & 92.38 & 92.95 & 92.79 & 92.79 & 92.87 & 93.35 $\pm$ 1.06 \\
      \midrule
      & AlexNet & 93.03 & 94.04 & 95.22 & \textbf{96.39} & 93.35 & 92.99 & 92.79 & 93.15 & 92.99 & 93.31 & 93.48 & 93.70 $\pm$ 1.07 \\
      \cmidrule{2-14}
      ModelNet40 - test & VGG-M & 93.64 & 95.91 & 96.07 & \textbf{97.37} & 94.12 & 93.80 & 94.08 & 94.25 & 93.68 & 94.41 & 94.17 & 94.68 $\pm$ 1.16 \\
      \cmidrule{2-14}
      & ResNet-50 & 93.76 & 96.64 & 95.91 & \textbf{96.92} & 94.08 & 93.68 & 94.45 & 94.29 & 94.45 & 94.29 & 94.00 & 94.77 $\pm$ 1.10 \\
      \toprule
      & AlexNet & 94.16 & \textbf{97.58} & 93.94 & 94.38 & 94.60 & 93.61 & 94.05 & 93.94 & 94.38 & 94.71 & 94.38 & 94.52 $\pm$ 1.01 \\
      \cmidrule{2-14}
      ModelNet10 - test & VGG-M & 94.38 & \textbf{98.46} & 94.05 & 94.93 & 94.38 & 94.38 & 94.38 & 94.49 & 94.82 & 94.49 & 94.27 & 94.82 $\pm$ 1.17 \\
      \cmidrule{2-14}
      & ResNet-50 & 94.82 & \textbf{97.80} & 94.49 & 94.49 & 94.16 & 94.60 & 94.38 & 94.60 & 94.27 & 94.71 & 94.49 & 94.80 $\pm$ 0.96 \\
      \bottomrule
    \end{tabular}
  \end{center}
  \caption{Comparison of classification accuracy (\%) with different camera system orientations. The best scores with each RotationNet architecture among the orientations are shown in bold.
}
  \label{table:camera_system_comparison}
  \normalsize
\end{table}

\section{Effectiveness of fine-tuning}
\label{sec:finetuning}

\begin{figure}[t]
  \begin{center}
    \begin{minipage}{.32\hsize}
      \includegraphics[width=\linewidth]{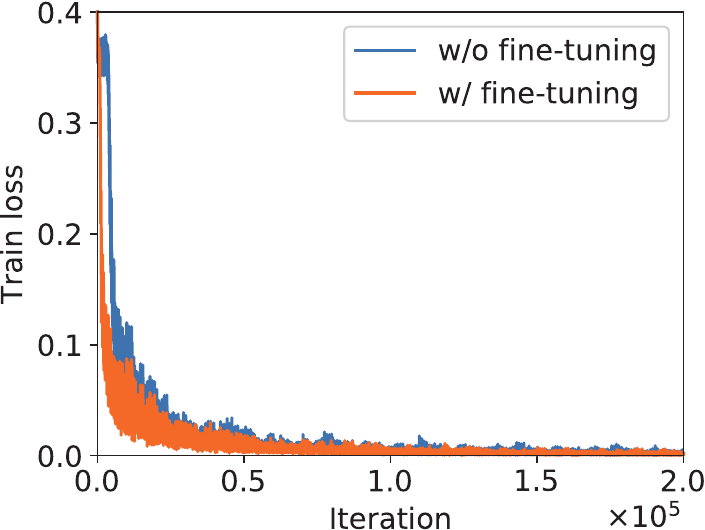}
      \centering
      \vspace{-1mm}
      \caption{Comparison of training loss on ModelNet40 using RotationNet trained w/ and w/o fine-tuning.}
      \label{fig:with_and_without_FT_loss}
    \end{minipage}      
    \hspace{1mm}
    \begin{minipage}{.32\hsize}
      \includegraphics[width=\linewidth]{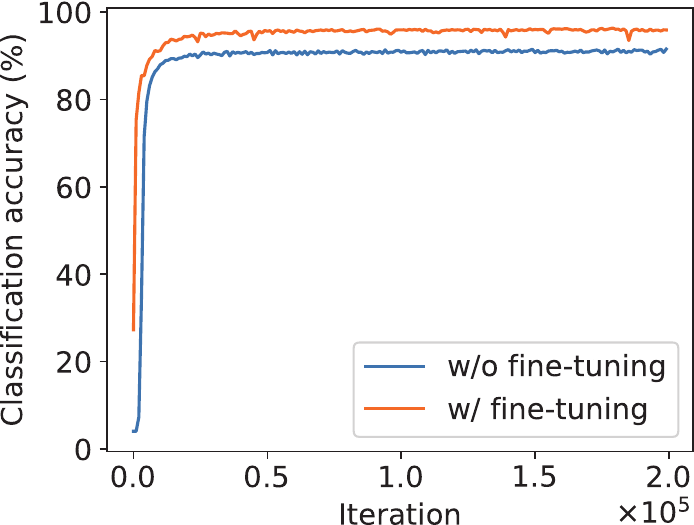}
      \centering
      \vspace{-1mm}
      \caption{Comparison of classification accuracy (\%) on ModelNet40 using RotationNet trained w/ and w/o fine-tuning.}
      \label{fig:with_and_without_FT_acc}
    \end{minipage}      
    \hspace{1mm}
    \begin{minipage}{.32\hsize}
      \includegraphics[width=\linewidth]{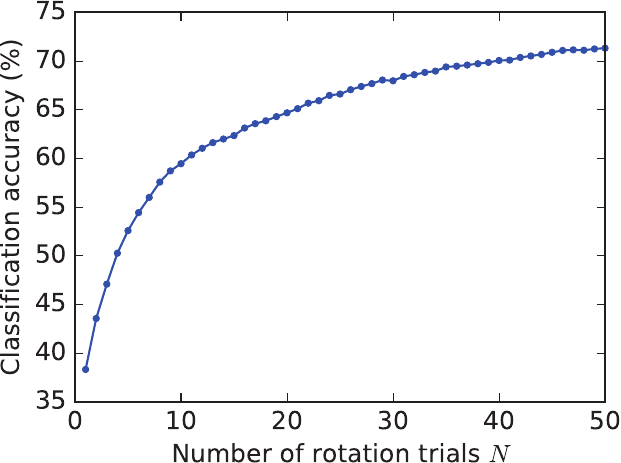}
      \centering
  \vspace{-1mm}
  \caption{Classification accuracy (\%) on ShapeNetCore55. We rotated a test model $N$ times and used the maximum scores.}
  \label{fig:sensitivity}
    \end{minipage}      
  \end{center}
\end{figure}

Even when the input images are grayscale rendered images of 3D models, fine-tuning of the ImageNet pre-trained weights is effective.
Figures \ref{fig:with_and_without_FT_loss} and \ref{fig:with_and_without_FT_acc} respectively show the training loss and the classification accuracy (\%) on ModelNet40 using RotationNet trained w/ and w/o fine-tuning.
Here, we used AlexNet as the baseline architecture of RotationNet.
As shown in these figures, fine-tuned RotationNet converges earlier to the optimal one that is better than the model achieved without fine-tuning.
It indicates that the ImageNet pre-trained weights capture general features of objects, which leads RotationNet to achieve reliable performance in the object classification task.

\section{Sensitivity to pre-defined views assumption}
\label{sec:sensitivity}

Similar to existing 3D object classification methods (such as MVCNN) that consider discrete variance of rotation, RotationNet has the limitation that each image should be observed from one of the pre-defined viewpoints.
To examine the sensitivity to pre-defined views assumption, we conducted an additional experiment with ShapeNetCore55 dataset\footnote{\url{http://shapenet.cs.stanford.edu/shrec16/}} which consists of 3D models in 55 object categories.
We trained our model with {\it aligned} training dataset and tested the classification of {\it unaligned} (\ie, randomly rotated) models.
The accuracy was $38\%$ whereas it was $90\%$ for aligned test models, which means our model is rather sensitive to pre-defined views.
However, as shown in \fref{fig:sensitivity}, the accuracy increases if we randomly rotate the test model $N$ times and use the maximum object scores.
Moreover, when trained with unaligned dataset (as is the case with ModelNet dataset), we achieve a model that is much less sensitive to the viewpoint sampling; the accuracy in this case was $73\%$ for unaligned test models with $N = 1$.

\section{Object instances in MIRO dataset}
\label{sec:MIROinstances}
\begin{figure}[h!]
  \begin{center}
    \includegraphics[width=\linewidth]{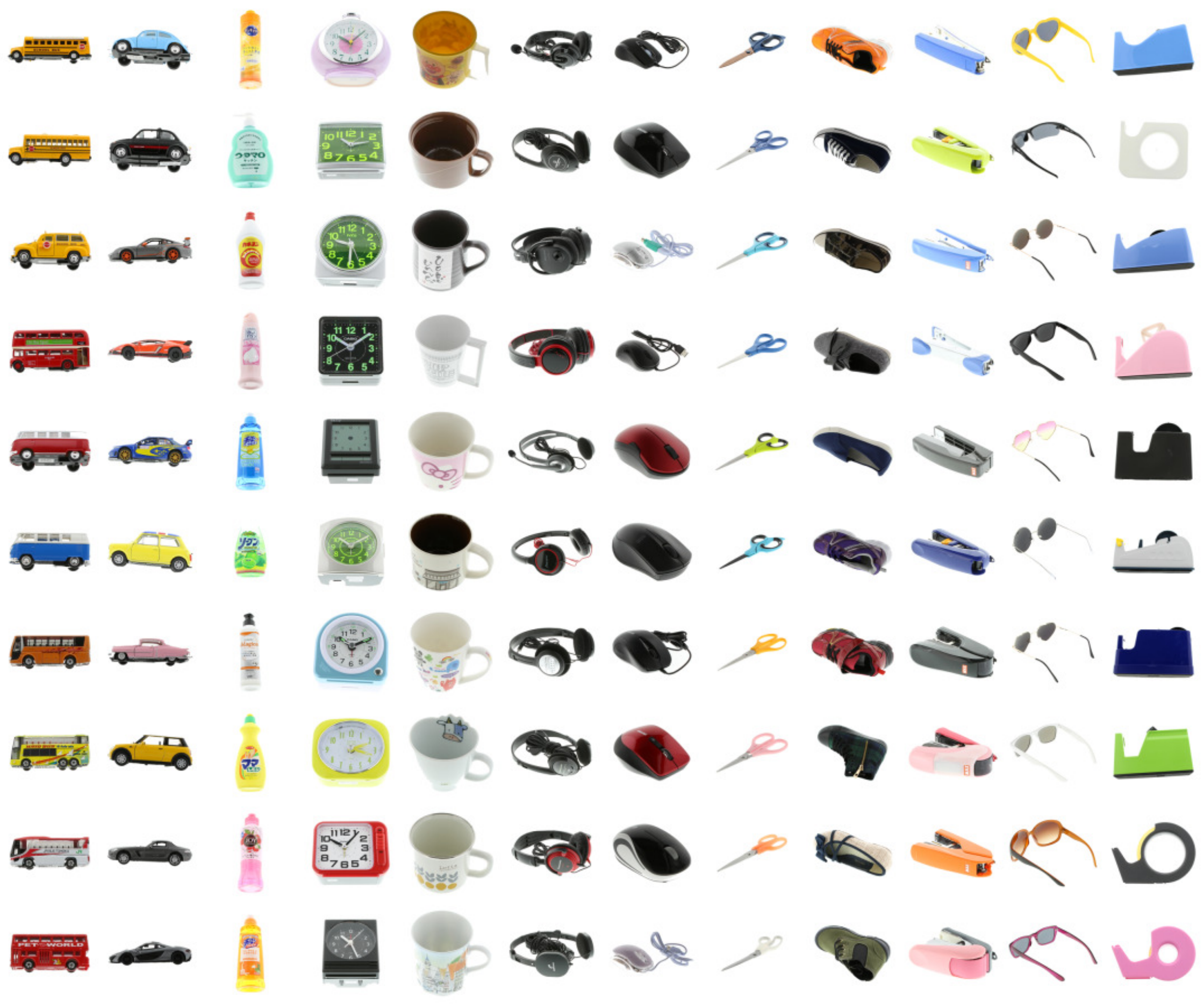}
  \end{center}
  \vspace{-3mm}
  \caption{Thumbnail images of all the object instances in MIRO. From left to right are shown the object instances in ``bus,'' ``car,'' ``cleanser,'' ``clock,'' ``cup,'' ``headphones,'' ``mouse,'' ``scissors,'' ``shoe,'' ``stapler,'' ``sunglasses,'' and ``tape\_cutter'' category.}
  \label{fig:thumbnail}
\end{figure}

\begin{figure}[h!]
  \begin{center}
    \includegraphics[width=\linewidth]{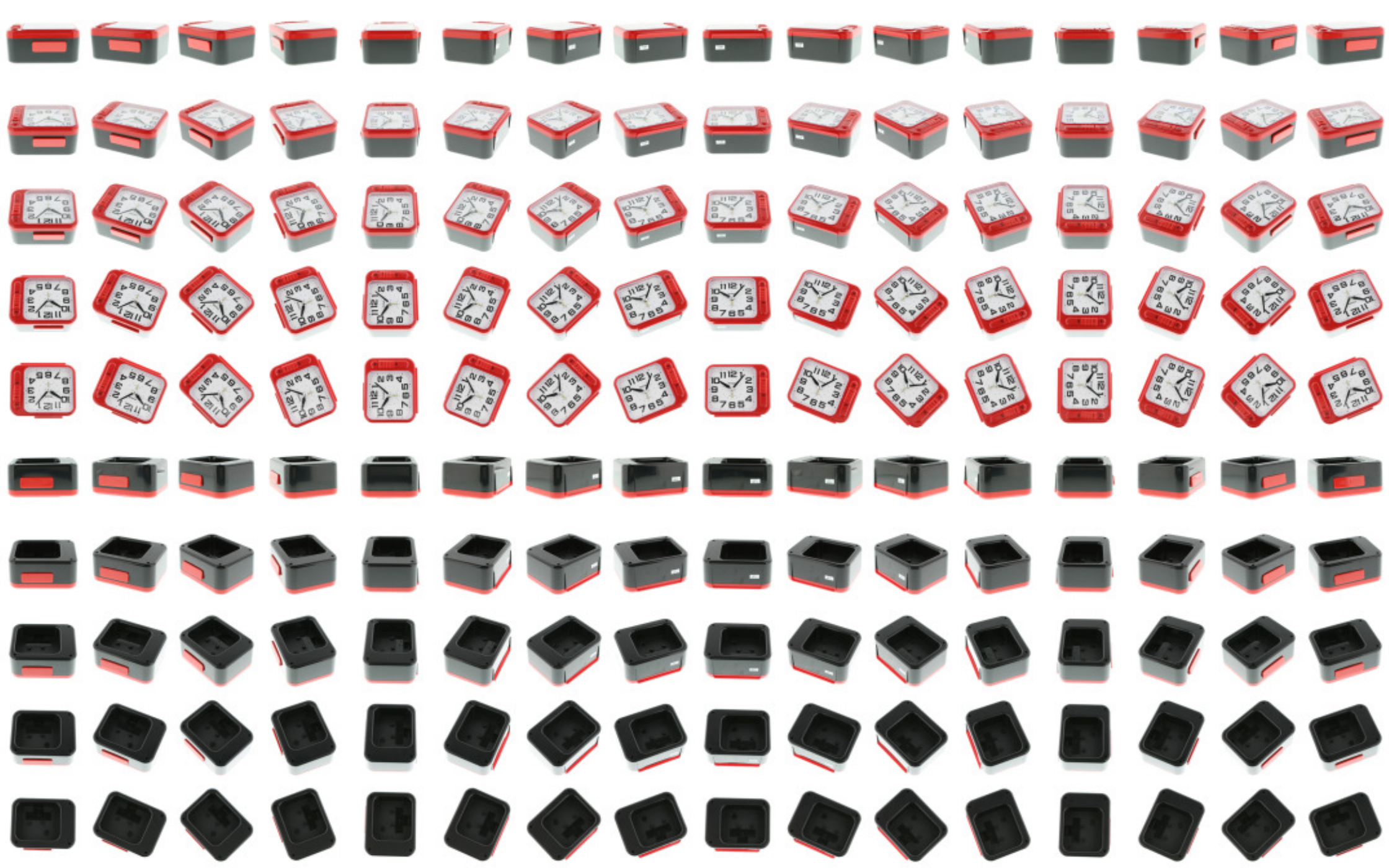}
  \end{center}
  \vspace{-3mm}
  \caption{All the $160$ images of an exemplar object instance captured from different viewpoints.}
  \label{fig:clock10_fig}
\end{figure}

Figure~\ref{fig:thumbnail} shows thumbnail images of all the object instances in our new dataset MIRO.
Our MIRO dataset includes $10$ object instances per object category. It consists of $120$ object instances in $12$ categories in total.
Each object instance has $160$ images captured from different viewpoints approximately equally distributed in the spherical coordinates. An example of the multi-view images are shown in Fig.~\ref{fig:clock10_fig}.

\clearpage
\section{Candidates for viewpoint variables $\{v_i\}_{i=1}^M$ in case (ii): w/o upright orientation}
\label{sec:candidates}
\begin{figure}[h]
  \begin{center}
    \includegraphics[width=\linewidth]{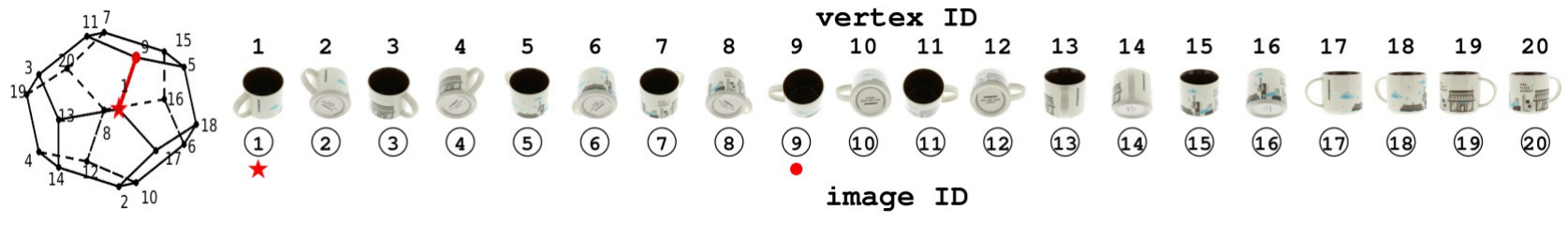}
    (a) Candidate \#1
    \includegraphics[width=\linewidth]{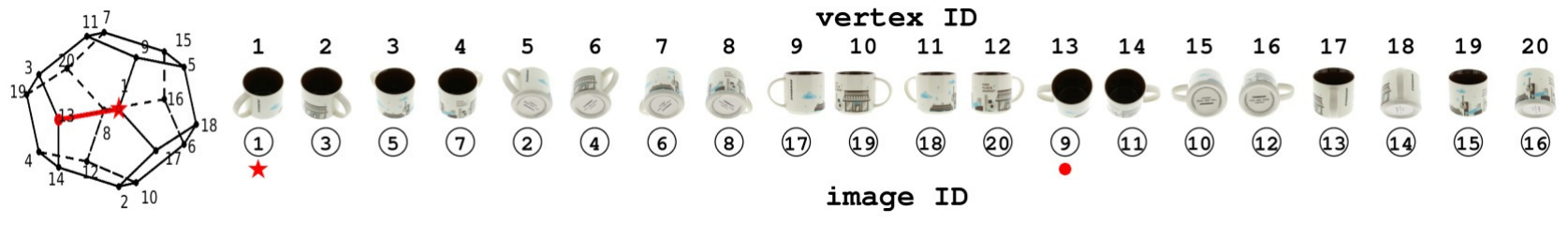}
    (b) Candidate \#2
    \includegraphics[width=\linewidth]{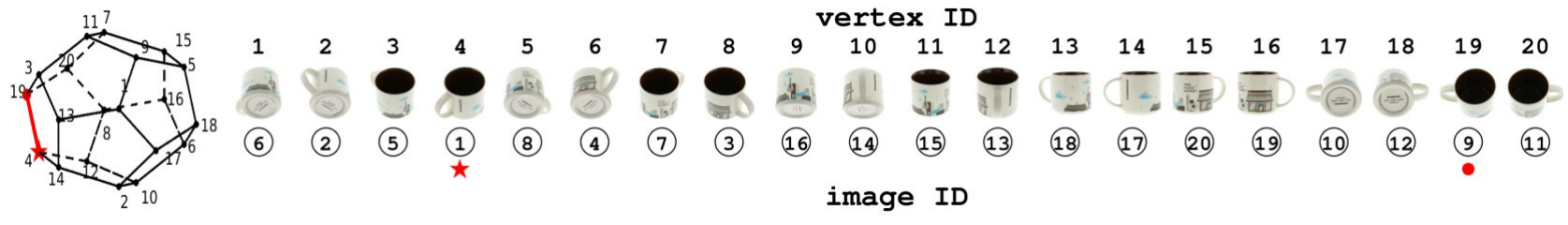}
    (c) Candidate \#3
    \includegraphics[width=\linewidth]{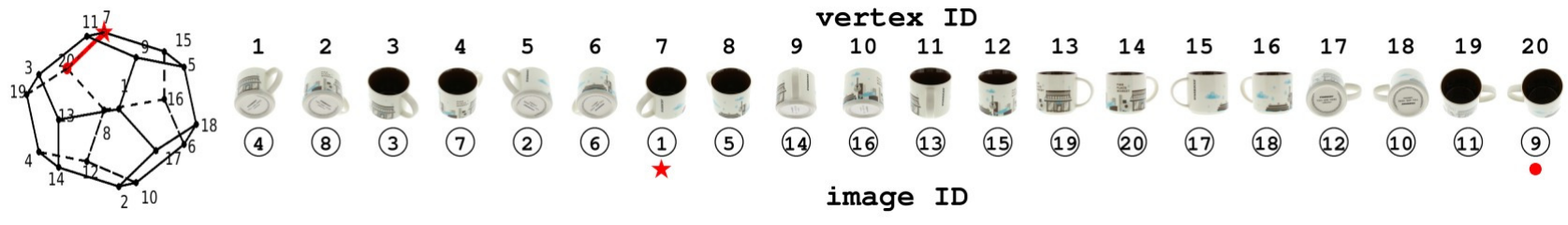}
    (d) Candidate \#4
  \end{center}
  \vspace{-2mm}
  \caption{Candidates \#1-\#4 for a set of viewpoint variables $\{v_i\}_{i=1}^{20}$ w/o upright orientation.}
  \label{fig:cand_1}
\end{figure}

\begin{figure}[t]
  \begin{center}
    \includegraphics[width=\linewidth]{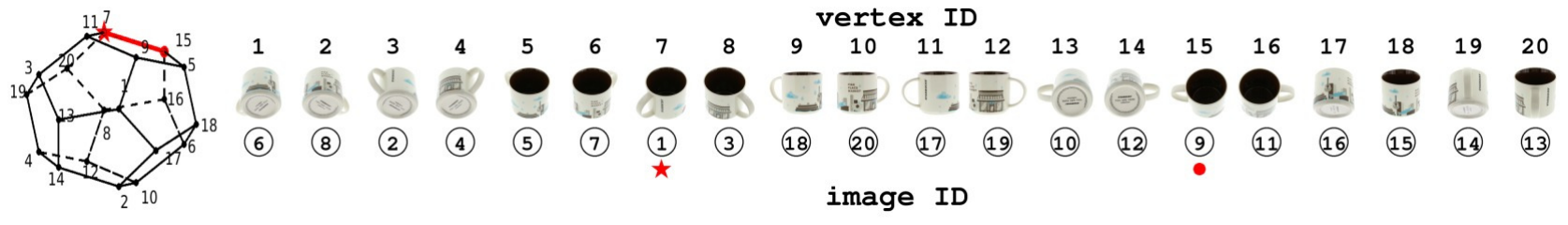}
    (a) Candidate \#5
    \includegraphics[width=\linewidth]{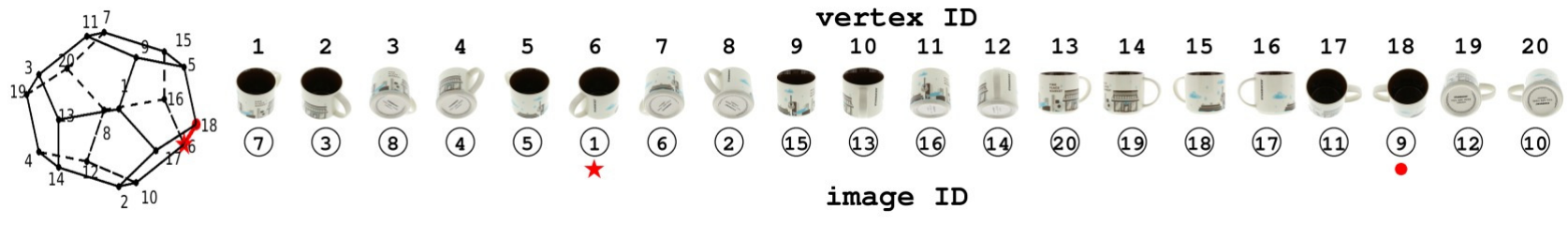}
    (b) Candidate \#6
    \includegraphics[width=\linewidth]{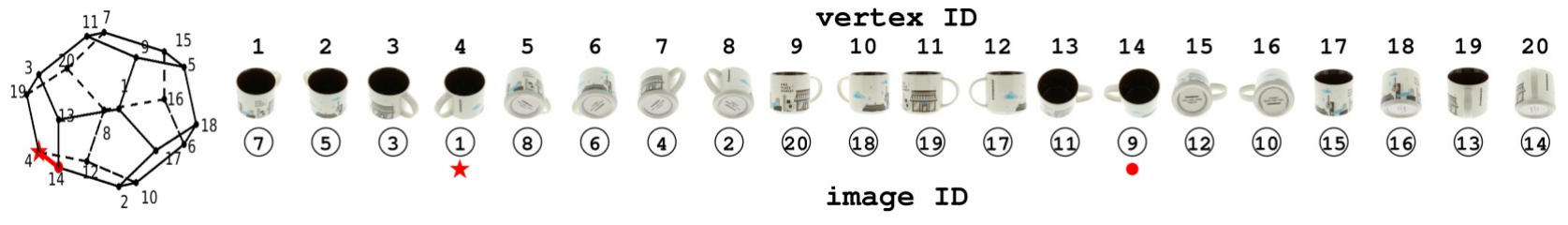}
    (c) Candidate \#7
    \includegraphics[width=\linewidth]{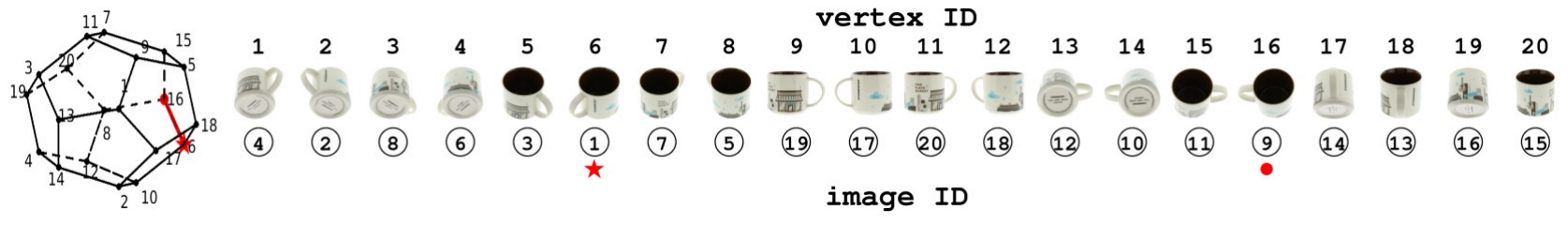}
    (d) Candidate \#8
    \includegraphics[width=\linewidth]{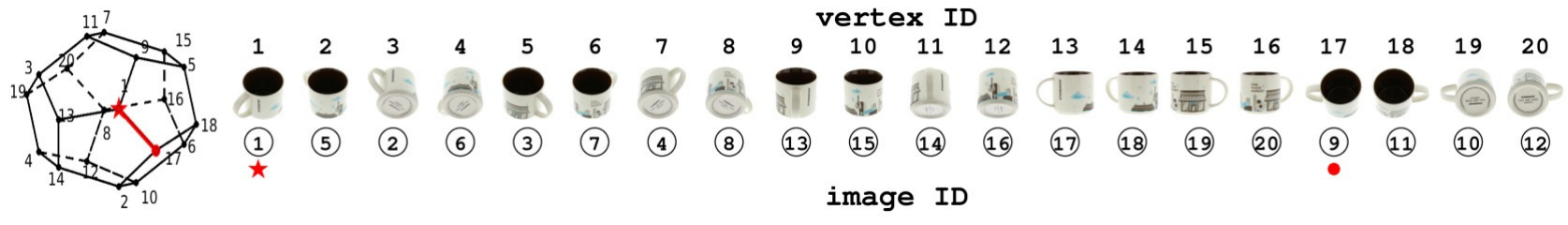}
    (e) Candidate \#9
    \includegraphics[width=\linewidth]{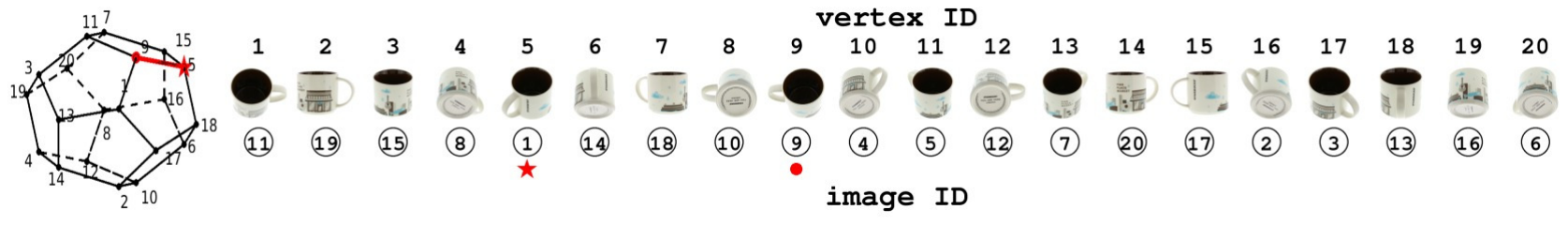}
    (f) Candidate \#10
    \includegraphics[width=\linewidth]{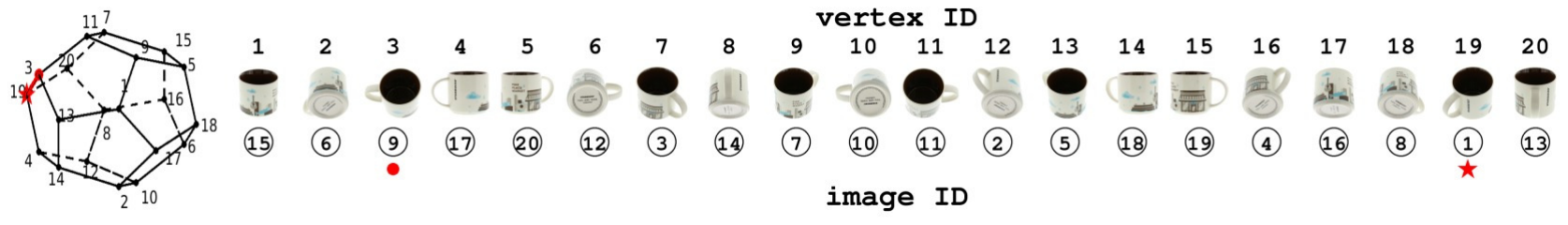}
    (g) Candidate \#11
  \end{center}
  \caption{Candidates \#5-\#11 for a set of viewpoint variables $\{v_i\}_{i=1}^{20}$ w/o upright orientation.}
  \label{fig:cand_2}
\end{figure}

\begin{figure}[t]
  \begin{center}
    \includegraphics[width=\linewidth]{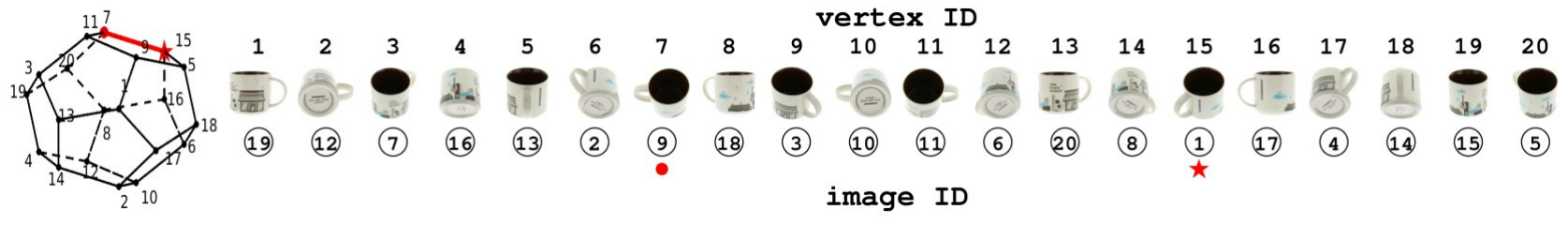}
    (a) Candidate \#12
    \includegraphics[width=\linewidth]{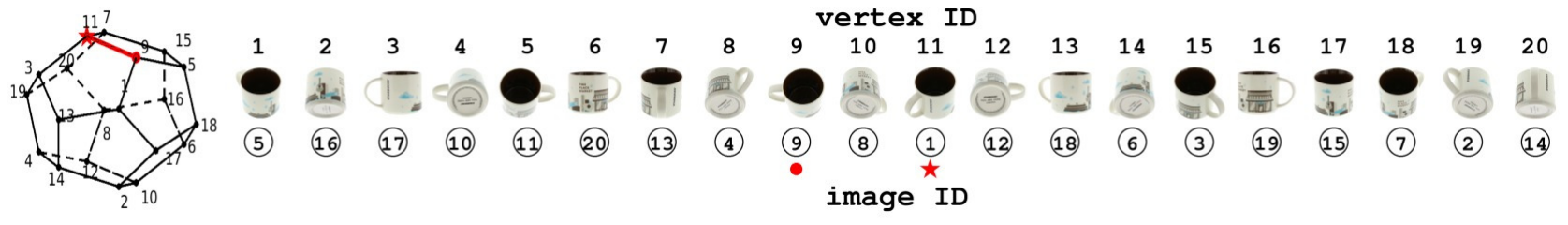}
    (b) Candidate \#13
    \includegraphics[width=\linewidth]{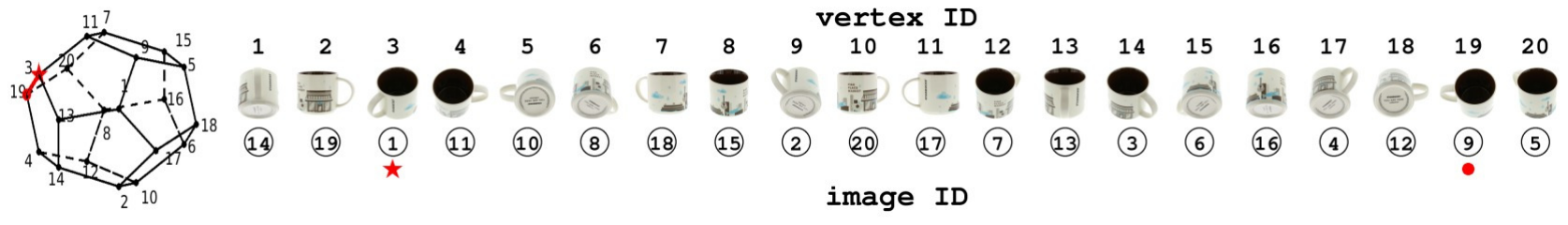}
    (c) Candidate \#14
    \includegraphics[width=\linewidth]{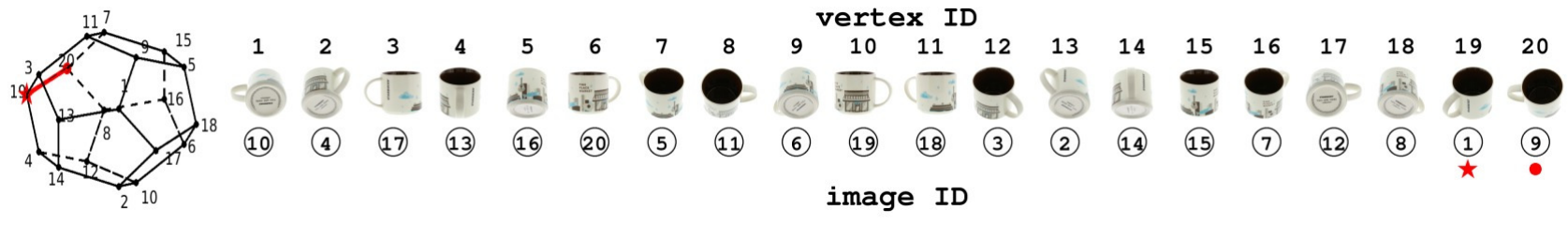}
    (d) Candidate \#15
    \includegraphics[width=\linewidth]{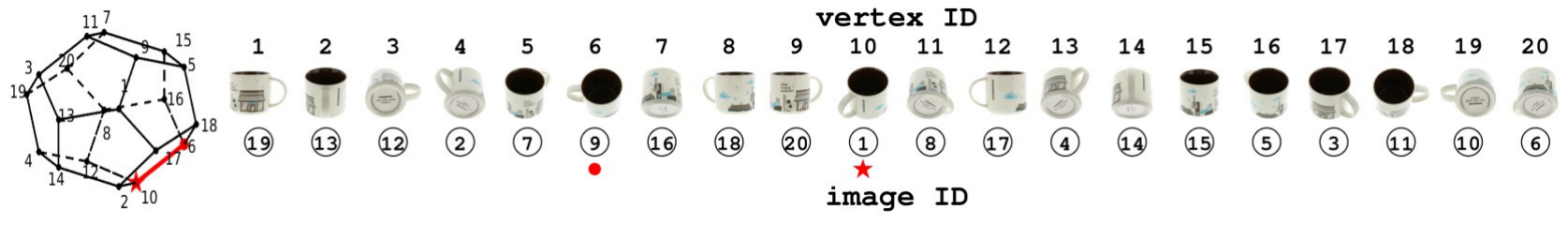}
    (e) Candidate \#16
    \includegraphics[width=\linewidth]{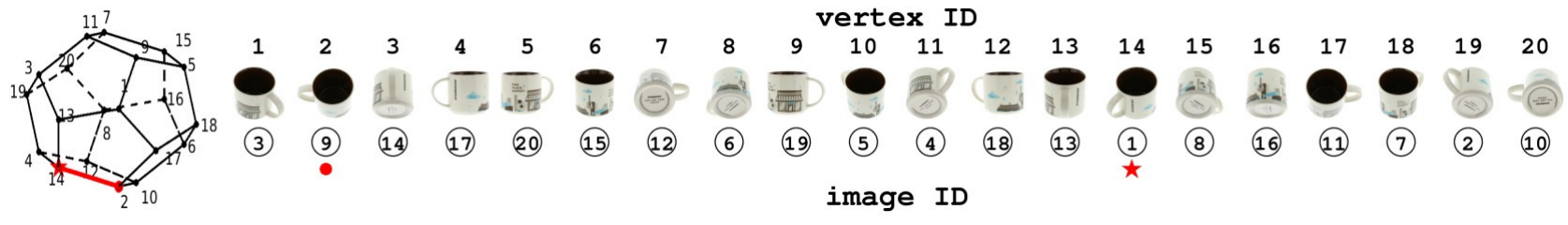}
    (f) Candidate \#17
    \includegraphics[width=\linewidth]{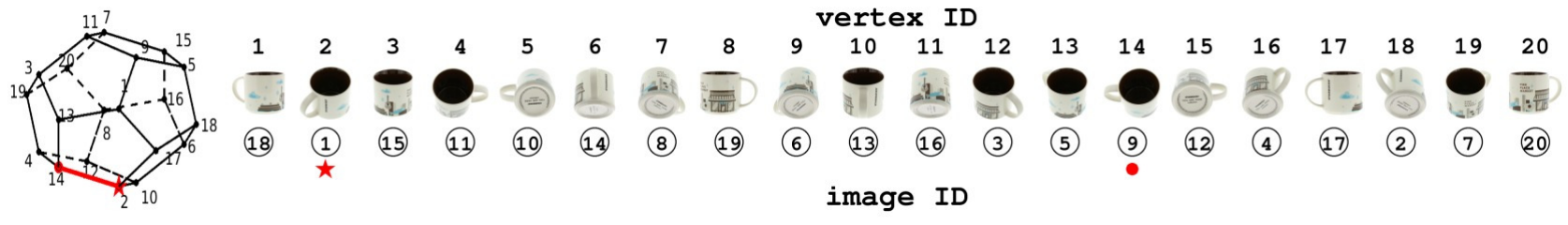}
    (g) Candidate \#18
  \end{center}
  \caption{Candidates \#12-\#18 for a set of viewpoint variables $\{v_i\}_{i=1}^{20}$ w/o upright orientation.}
  \label{fig:cand_3}
\end{figure}

\begin{figure}[t]
  \begin{center}
    \includegraphics[width=\linewidth]{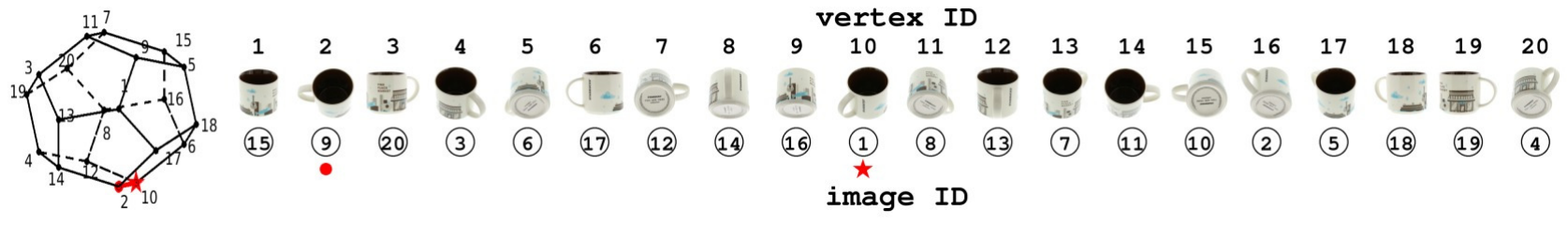}
    (a) Candidate \#19
    \includegraphics[width=\linewidth]{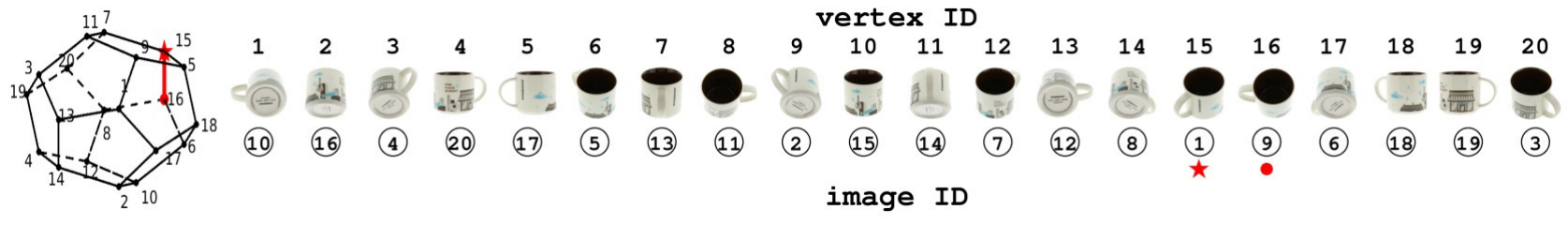}
    (b) Candidate \#20
    \includegraphics[width=\linewidth]{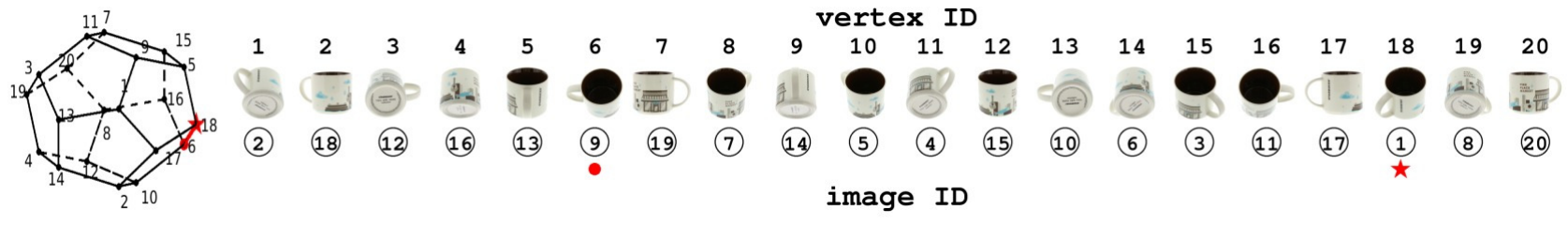}
    (c) Candidate \#21
    \includegraphics[width=\linewidth]{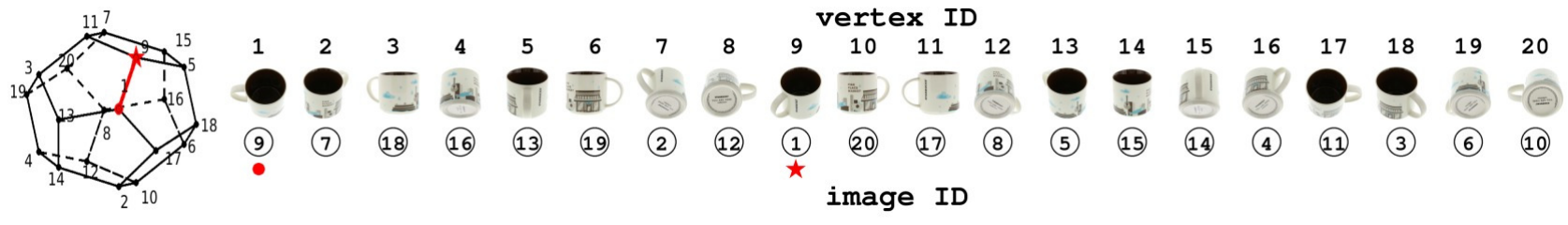}
    (d) Candidate \#22
    \includegraphics[width=\linewidth]{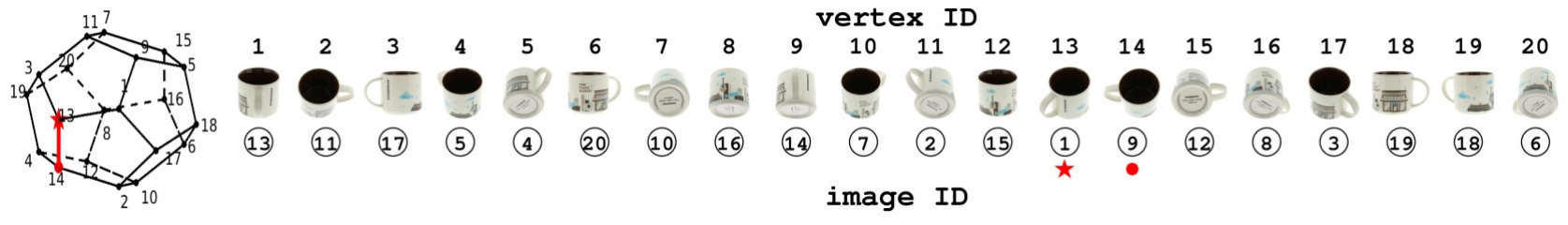}
    (e) Candidate \#23
    \includegraphics[width=\linewidth]{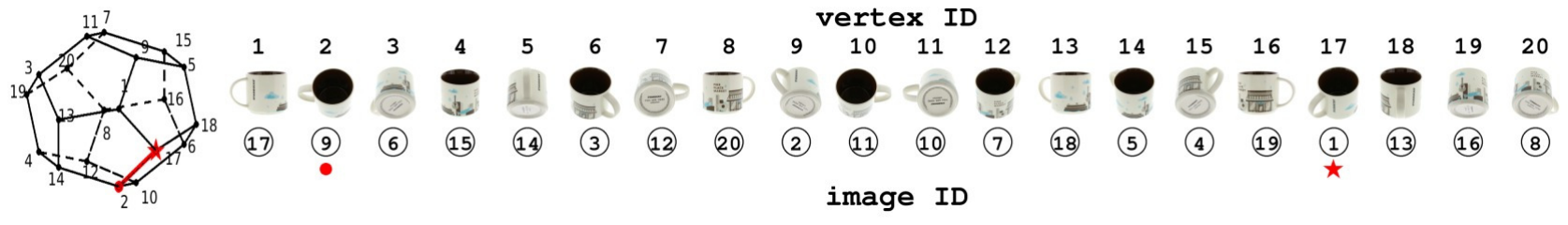}
    (f) Candidate \#24
    \includegraphics[width=\linewidth]{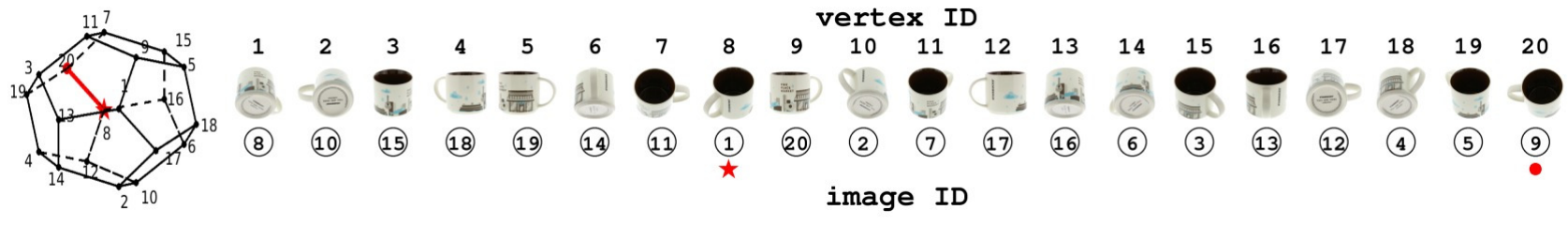}
    (g) Candidate \#25
  \end{center}
  \caption{Candidates \#19-\#25 for a set of viewpoint variables $\{v_i\}_{i=1}^{20}$ w/o upright orientation.}
  \label{fig:cand_4}
\end{figure}

\begin{figure}[t]
  \begin{center}
    \includegraphics[width=\linewidth]{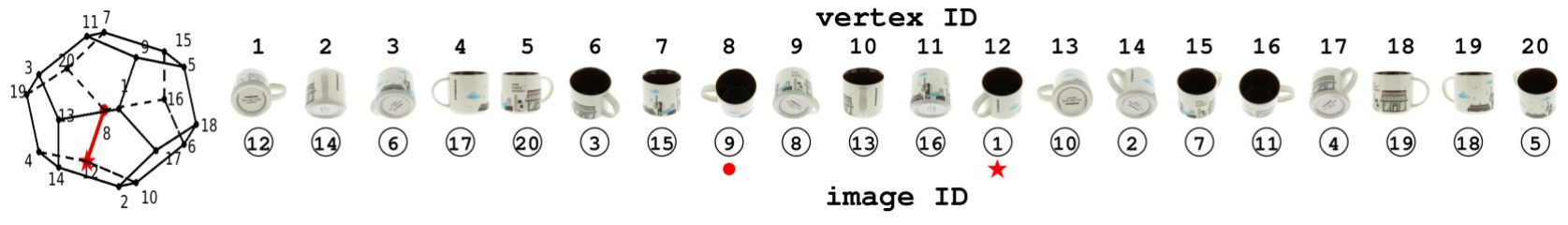}
    (a) Candidate \#26
    \includegraphics[width=\linewidth]{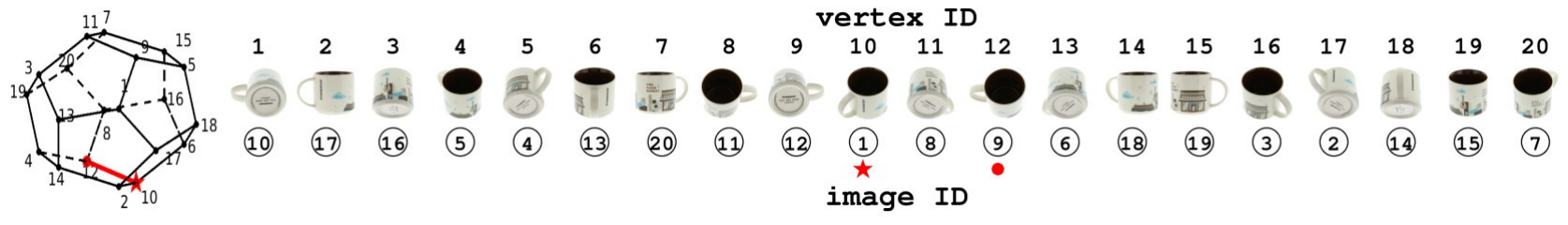}
    (b) Candidate \#27
    \includegraphics[width=\linewidth]{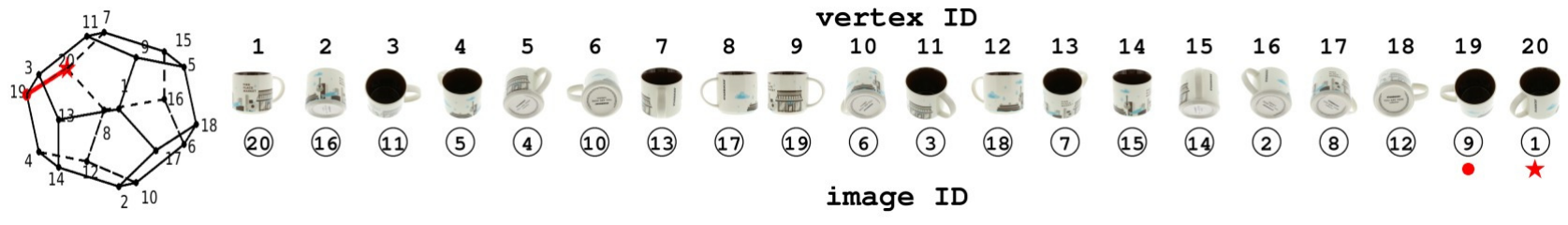}
    (c) Candidate \#28
    \includegraphics[width=\linewidth]{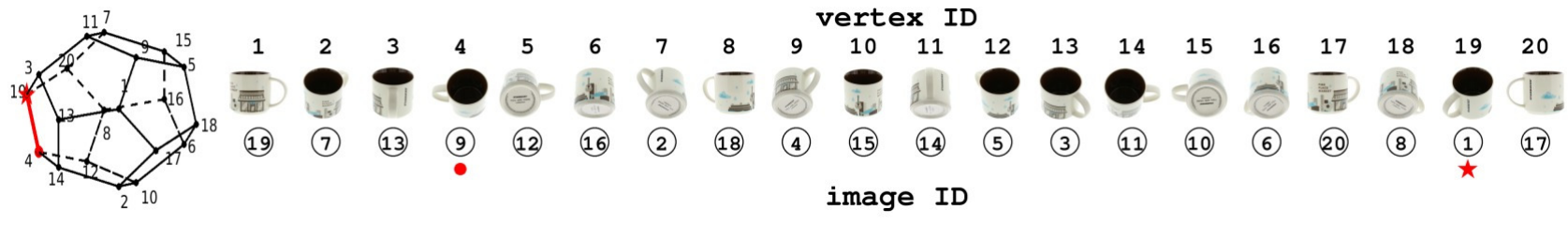}
    (d) Candidate \#29
    \includegraphics[width=\linewidth]{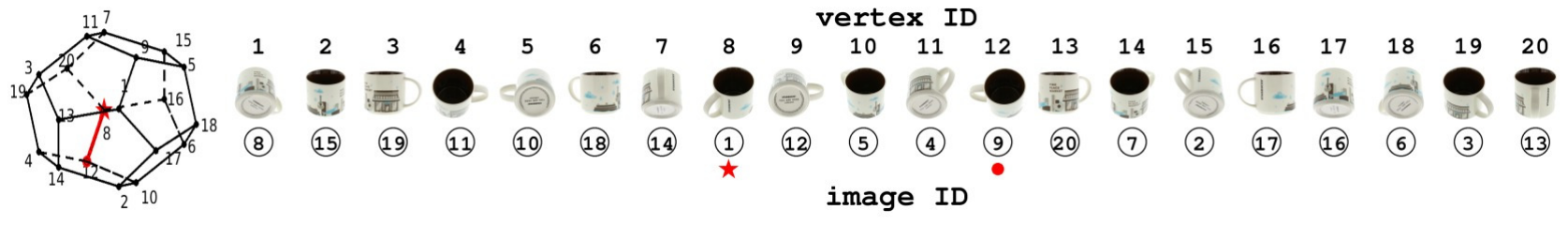}
    (e) Candidate \#30
    \includegraphics[width=\linewidth]{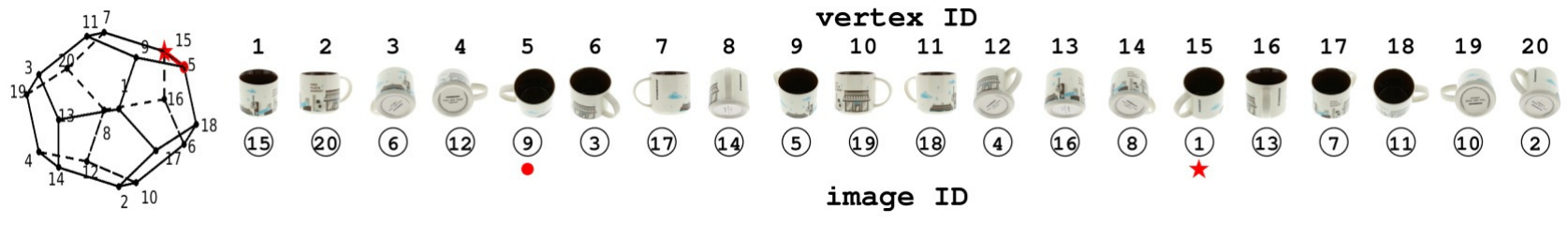}
    (f) Candidate \#31
    \includegraphics[width=\linewidth]{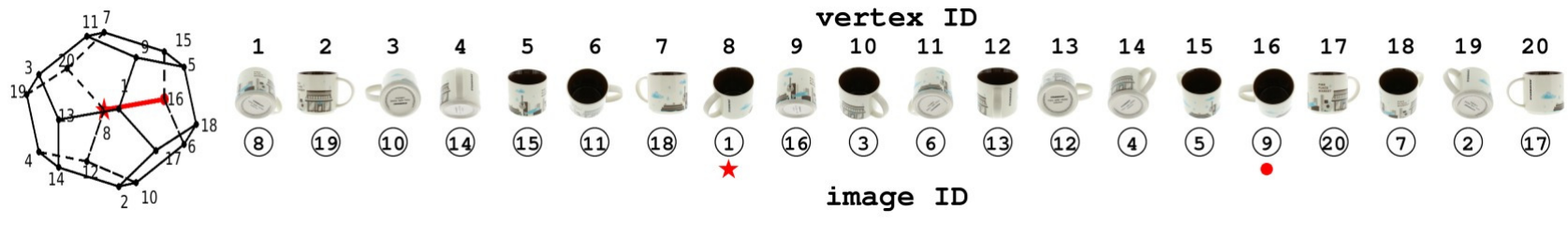}
    (g) Candidate \#32
  \end{center}
  \caption{Candidates \#26-\#32 for a set of viewpoint variables $\{v_i\}_{i=1}^{20}$ w/o upright orientation.}
  \label{fig:cand_5}
\end{figure}

\begin{figure}[t]
  \begin{center}
    \includegraphics[width=\linewidth]{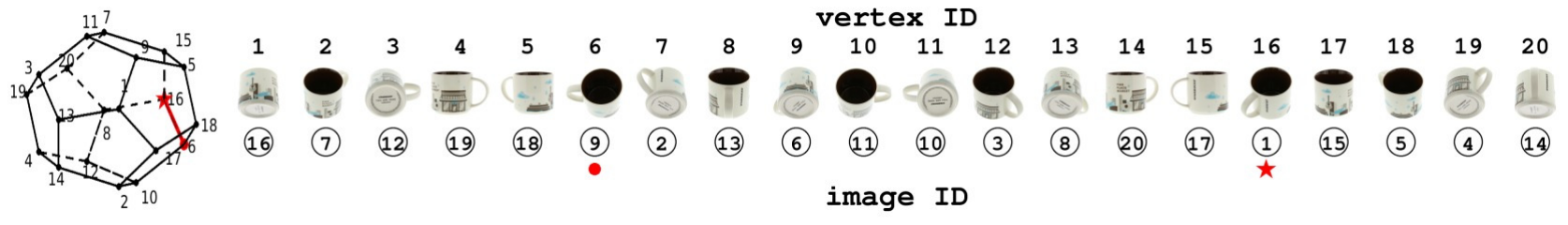}
    (a) Candidate \#33
    \includegraphics[width=\linewidth]{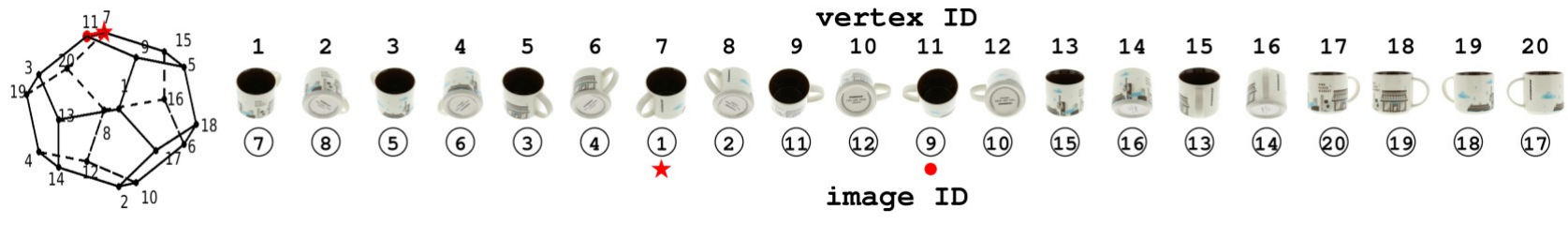}
    (b) Candidate \#34
    \includegraphics[width=\linewidth]{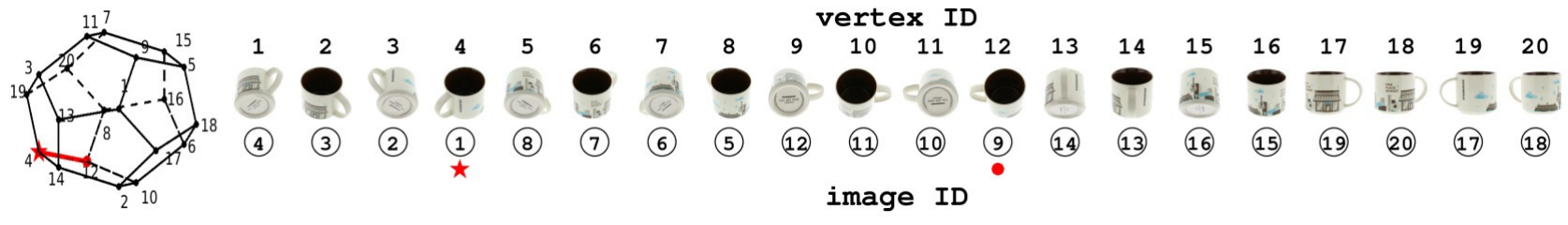}
    (c) Candidate \#35
    \includegraphics[width=\linewidth]{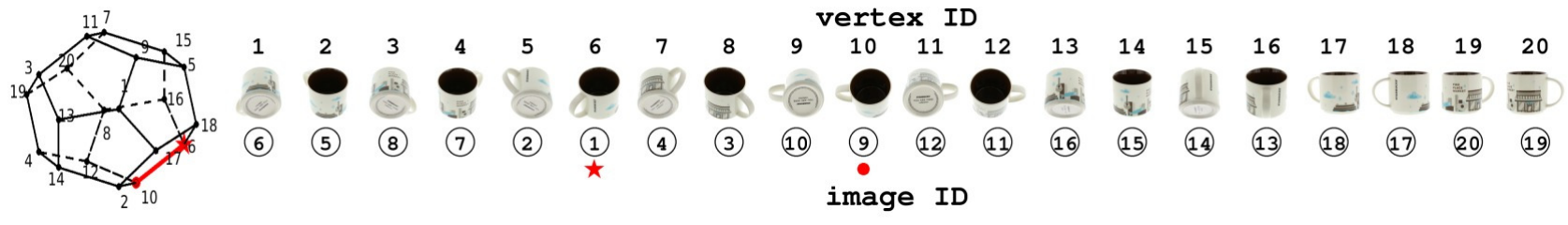}
    (d) Candidate \#36
    \includegraphics[width=\linewidth]{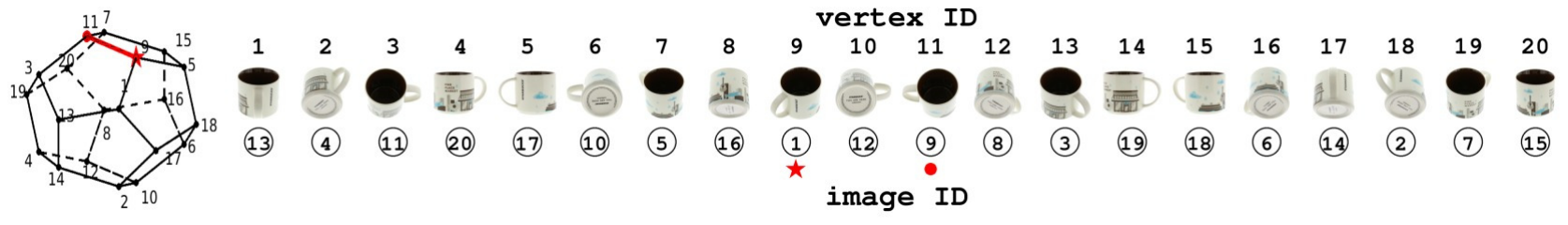}
    (e) Candidate \#37
    \includegraphics[width=\linewidth]{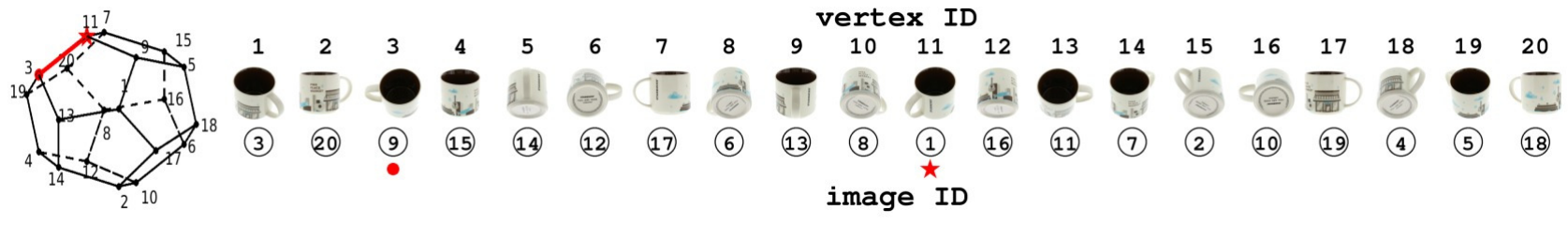}
    (f) Candidate \#38
    \includegraphics[width=\linewidth]{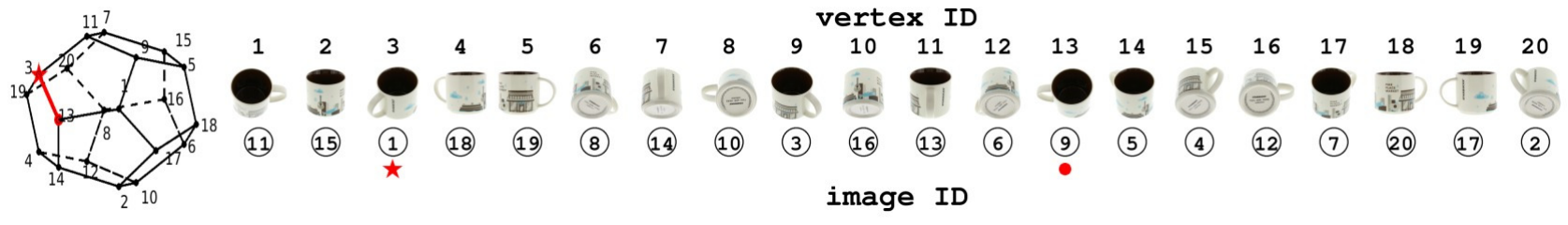}
    (g) Candidate \#39
  \end{center}
  \caption{Candidates \#33-\#39 for a set of viewpoint variables $\{v_i\}_{i=1}^{20}$ w/o upright orientation.}
  \label{fig:cand_6}
\end{figure}

\begin{figure}[t]
  \begin{center}
    \includegraphics[width=\linewidth]{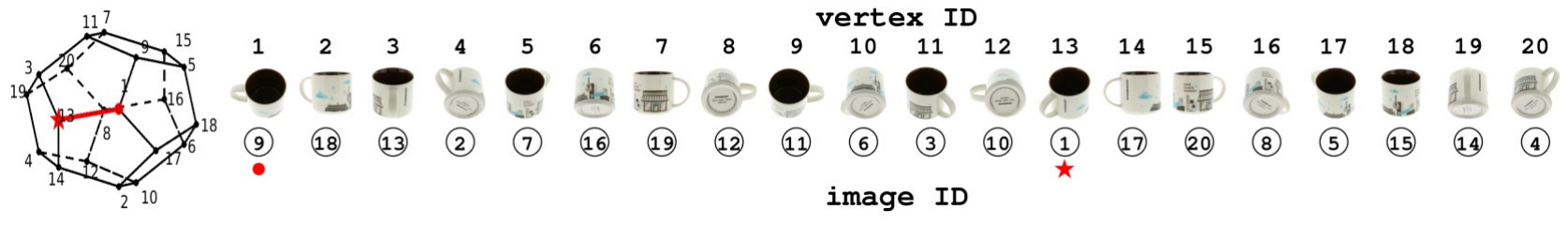}
    (a) Candidate \#40
    \includegraphics[width=\linewidth]{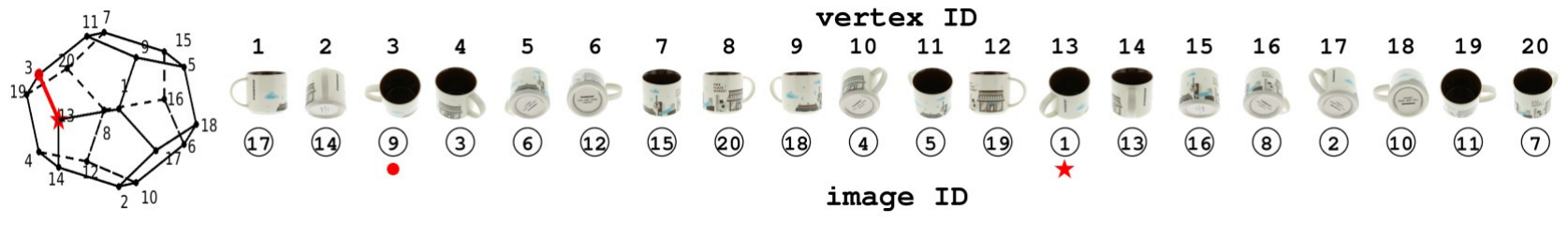}
    (b) Candidate \#41
    \includegraphics[width=\linewidth]{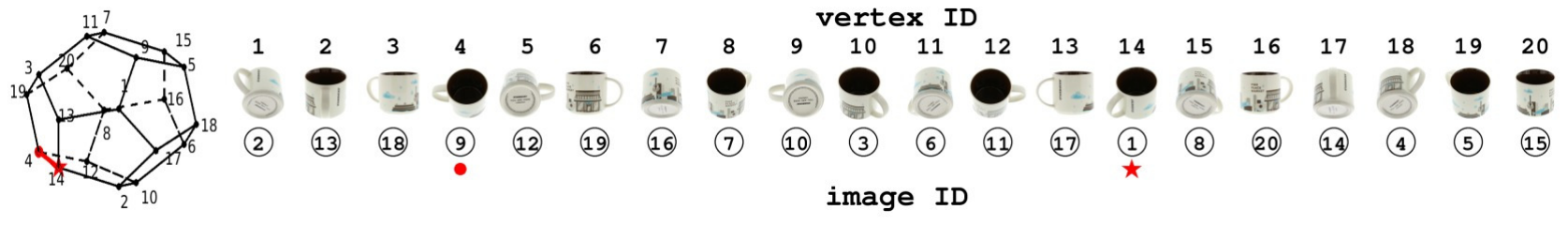}
    (c) Candidate \#42
    \includegraphics[width=\linewidth]{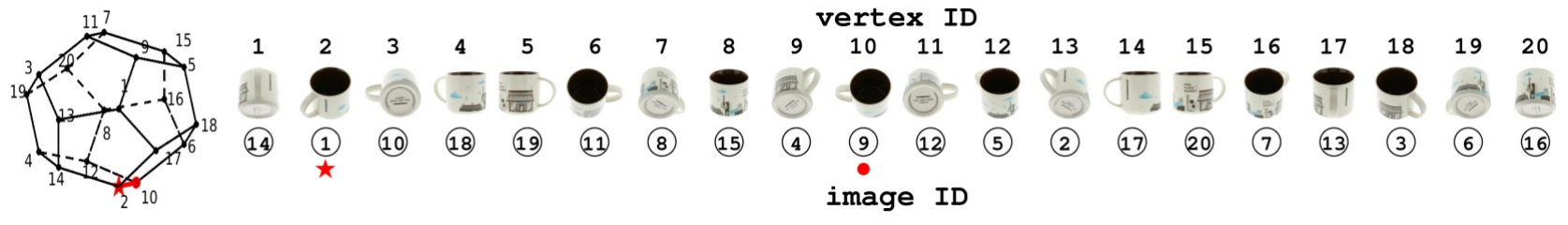}
    (d) Candidate \#43
    \includegraphics[width=\linewidth]{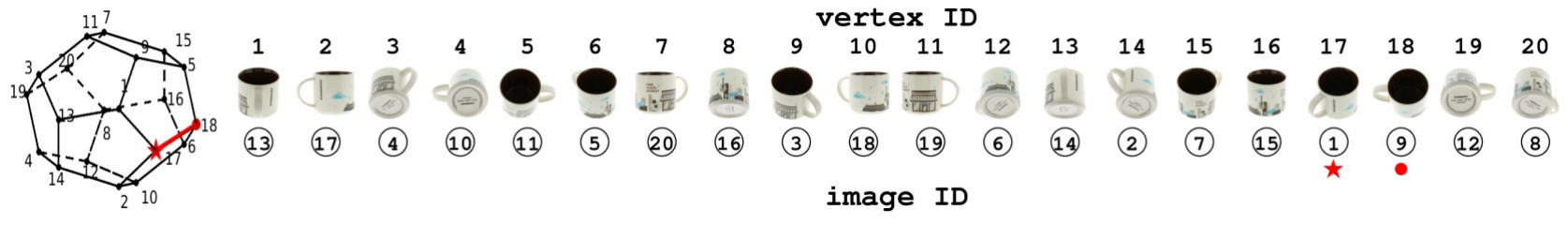}
    (e) Candidate \#44
    \includegraphics[width=\linewidth]{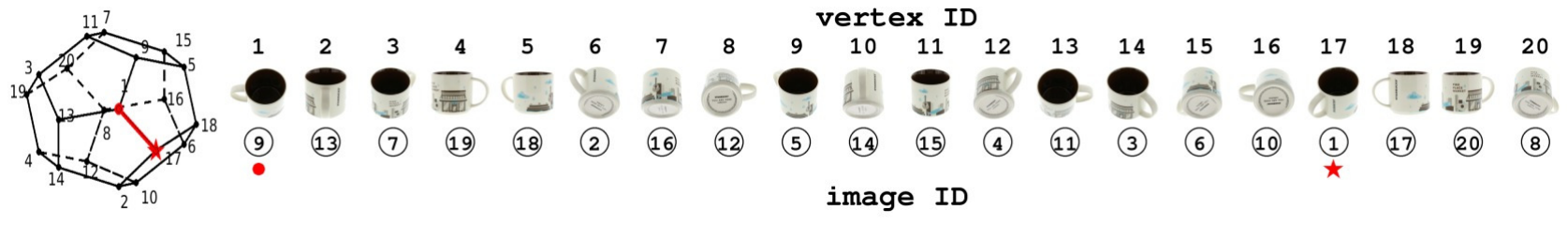}
    (f) Candidate \#45
    \includegraphics[width=\linewidth]{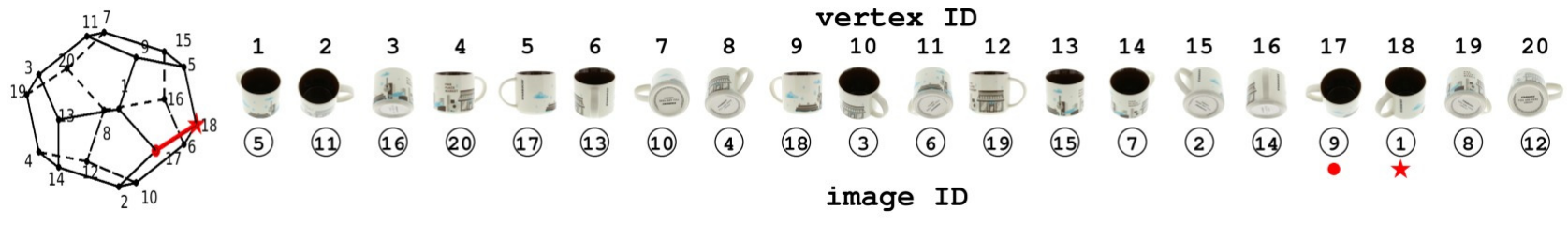}
    (g) Candidate \#46
  \end{center}
  \caption{Candidates \#40-\#46 for a set of viewpoint variables $\{v_i\}_{i=1}^{20}$ w/o upright orientation.}
  \label{fig:cand_7}
\end{figure}

\begin{figure}[t]
  \begin{center}
    \includegraphics[width=\linewidth]{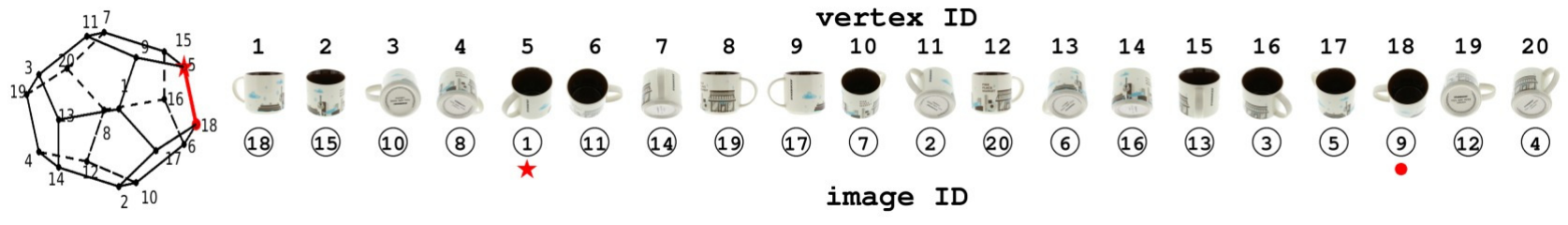}
    (a) Candidate \#47
    \includegraphics[width=\linewidth]{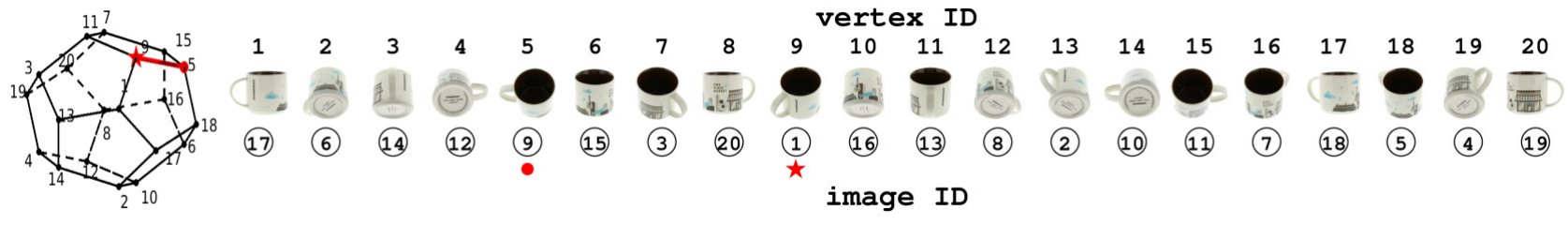}
    (b) Candidate \#48
    \includegraphics[width=\linewidth]{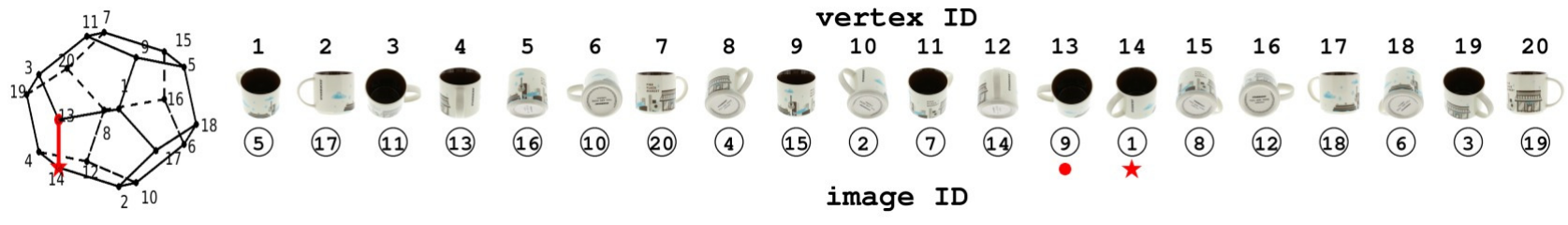}
    (c) Candidate \#49
    \includegraphics[width=\linewidth]{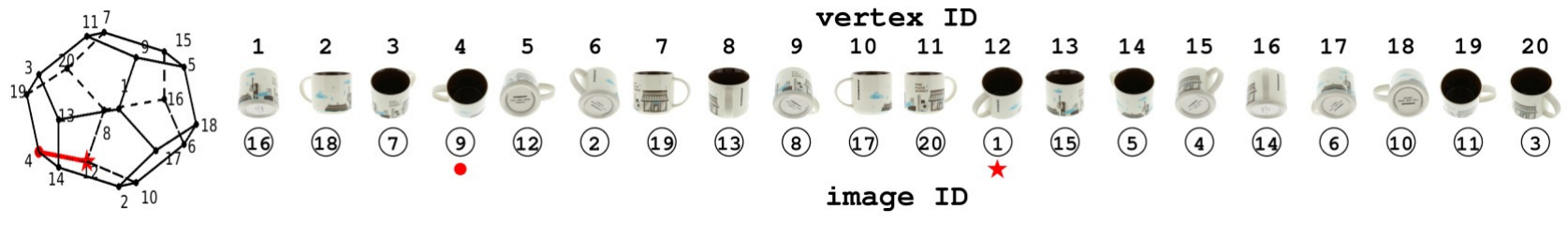}
    (d) Candidate \#50
    \includegraphics[width=\linewidth]{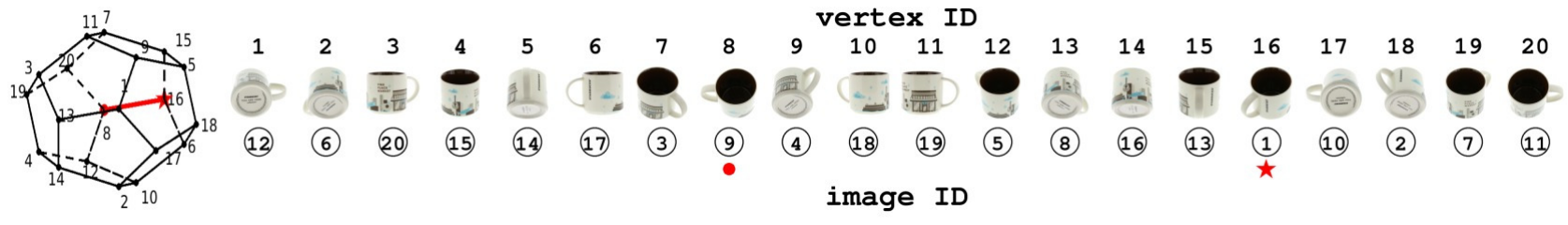}
    (e) Candidate \#51
    \includegraphics[width=\linewidth]{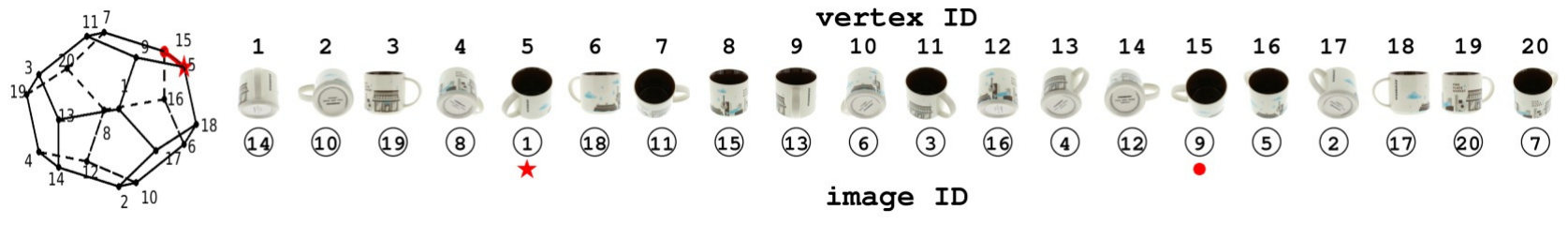}
    (f) Candidate \#52
    \includegraphics[width=\linewidth]{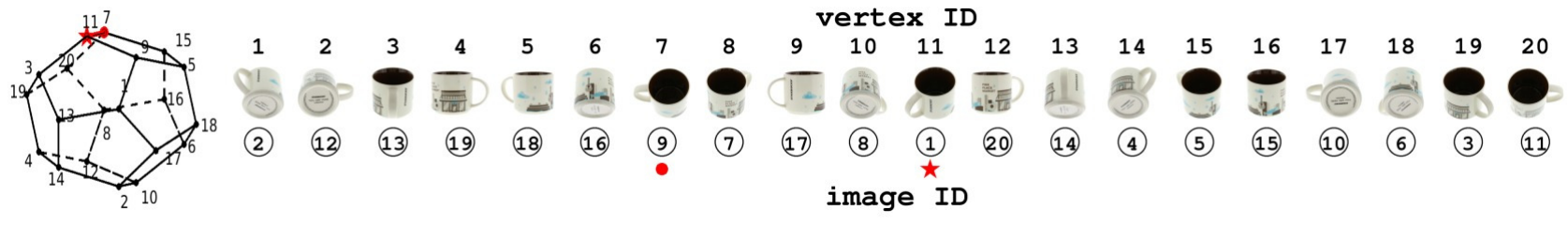}
    (g) Candidate \#53
  \end{center}
  \caption{Candidates \#47-\#53 for a set of viewpoint variables $\{v_i\}_{i=1}^{20}$ w/o upright orientation.}
  \label{fig:cand_8}
\end{figure}

\begin{figure}[t]
  \begin{center}
    \includegraphics[width=\linewidth]{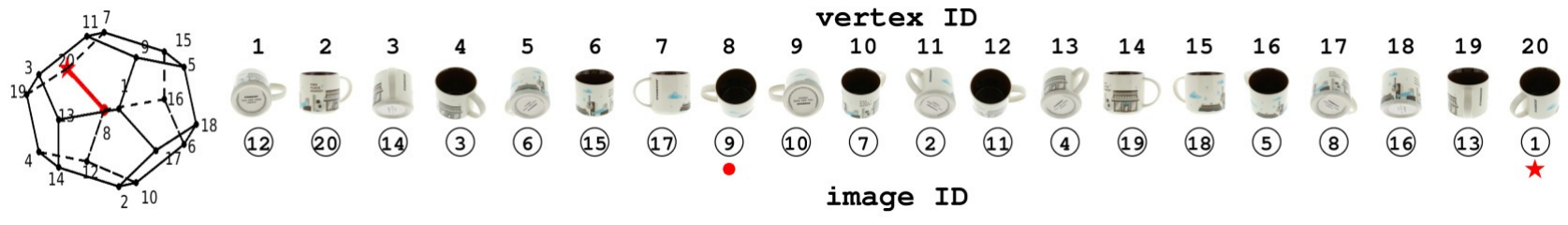}
    (a) Candidate \#54
    \includegraphics[width=\linewidth]{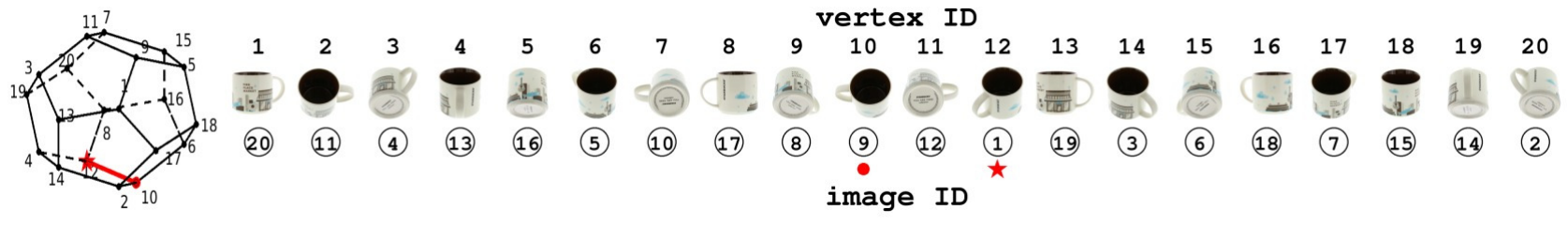}
    (b) Candidate \#55
    \includegraphics[width=\linewidth]{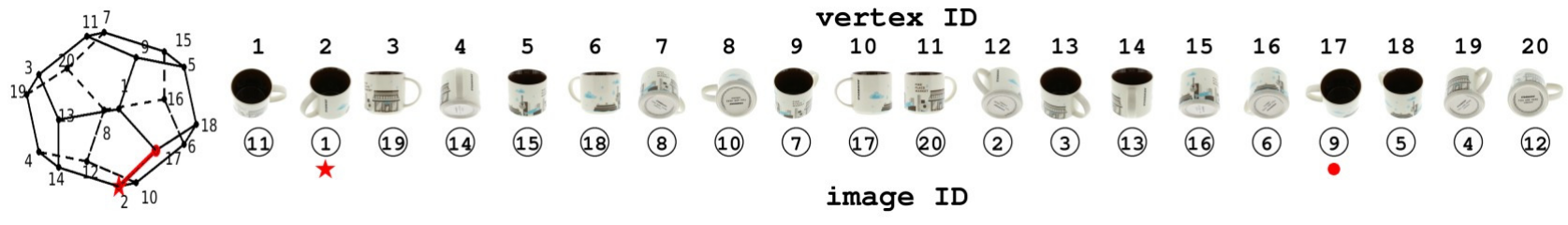}
    (c) Candidate \#56
    \includegraphics[width=\linewidth]{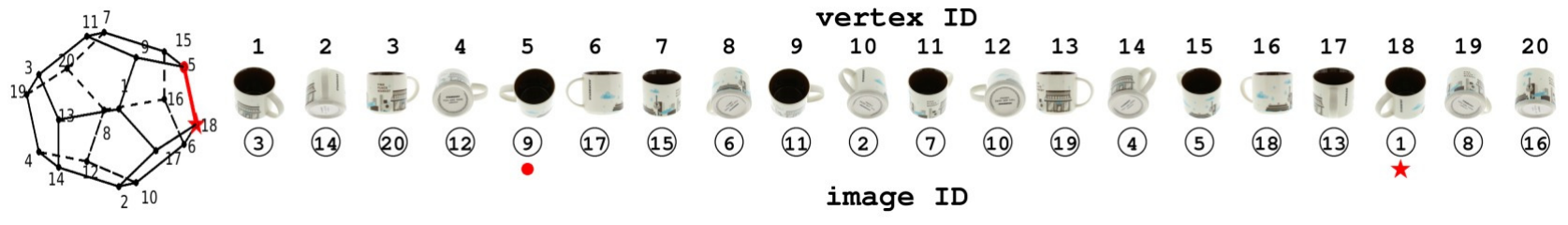}
    (d) Candidate \#57
    \includegraphics[width=\linewidth]{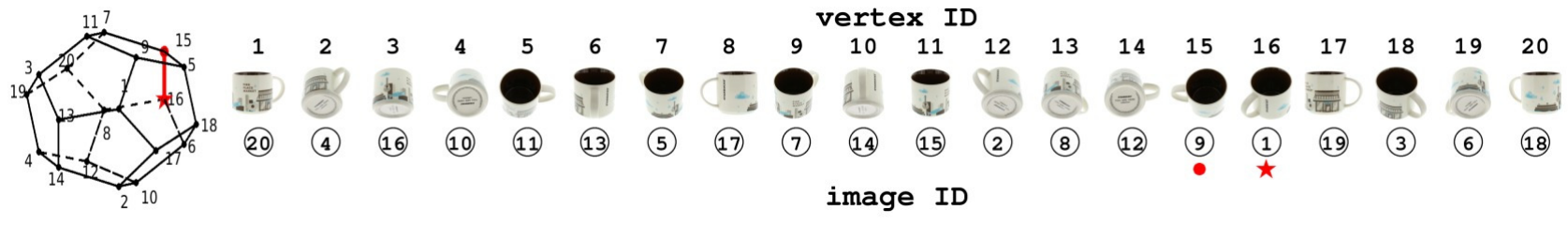}
    (e) Candidate \#58
    \includegraphics[width=\linewidth]{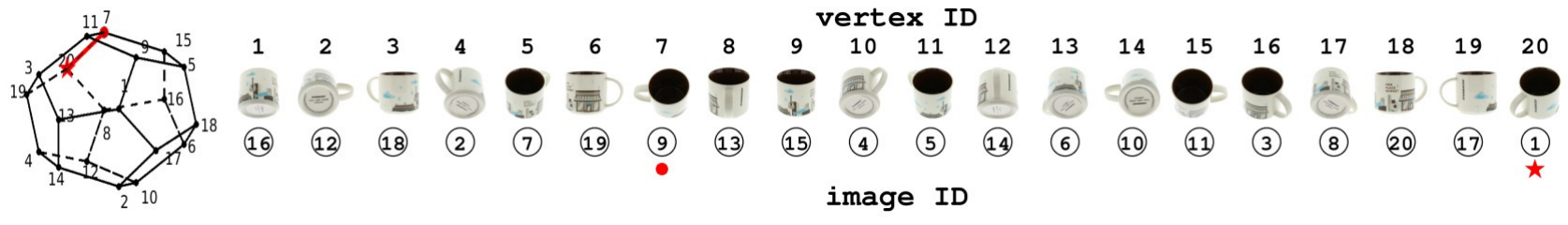}
    (f) Candidate \#59
    \includegraphics[width=\linewidth]{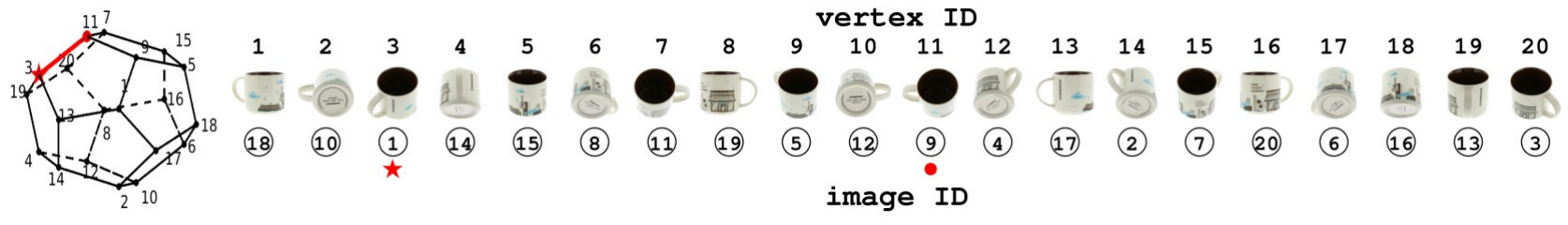}
    (g) Candidate \#60
  \end{center}
  \caption{Candidates \#54-\#60 for a set of viewpoint variables $\{v_i\}_{i=1}^{20}$ w/o upright orientation.}
  \label{fig:cand_9}
\end{figure}

In the case where we do not assume upright orientation, we place virtual cameras on the $M=20$ vertices of a dodecahedron encompassing the object.
There are three different patterns of rotation from a certain view, because three edges are connected to each vertex of a dodecahedron.
Therefore, the number of candidates for all the viewpoint variables $\{v_i\}_{i=1}^M$ is $60$ ($= 3M$).
Figures~\ref{fig:cand_1}-\ref{fig:cand_9} show all the candidates for a set of viewpoint variables $\{v_i\}_{i=1}^{20}$ in this case, in which vertex and image IDs are shown on the top and bottom rows respectively.
Here, $v_i$ indicates the ID of the vertex where the $i$-th image of the object instance is observed.
For instance, $\{v_i\}_{i=1}^{20}$ in Candidate \#2 is 
$\{1, 5, 2, 6, 3, 7, 4, 8, 13, 15, 14, 16, 17, 18, 19, 20, 9, 11, 10, 12\}$ (Fig.~\ref{fig:cand_1} (b)).
The red star indicates the camera position where the first image is captured. The red dot indicates the camera position where the ninth image is captured.

\end{document}